\documentclass[10pt,twocolumn,letterpaper]{article}

\usepackage[pagenumbers]{cvpr} 
\usepackage{lipsum}
\usepackage{amsmath}
\usepackage{amsthm}
\usepackage{acronym}
\usepackage{algpseudocode}
\usepackage{algorithm}
\usepackage{xcolor}
\usepackage{amsmath}
\usepackage{amssymb}
\usepackage{array}
\usepackage{multirow}
\usepackage{color}
\usepackage{colortbl}
\usepackage{framed}
\usepackage{bm}
\usepackage{bbm}
\usepackage{xspace}
\usepackage{enumitem}
\usepackage{algorithm}
\usepackage{listings}
\usepackage{url}
\usepackage{booktabs}
\usepackage{pifont} 
\usepackage[accsupp]{axessibility}
\usepackage{lipsum}
\usepackage{bm}
\usepackage[table]{xcolor}
\definecolor{LightCyan}{rgb}{0.88,0.95,1}
\definecolor{blond}{rgb}{0.98, 0.94, 0.75}
\definecolor{green}{rgb}{0.0, 0.62, 0.42}
\definecolor{LightGray}{gray}{0.93}
\definecolor{ourcolor}{rgb}{0.925, 0.902, 0.969}
\definecolor{ourcolorl}{rgb}{0.95, 0.945, 0.975}

\definecolor{myPink}{RGB}{247, 219, 231}  
\definecolor{myGreen}{RGB}{227, 241, 217}   
\definecolor{myBlue}{RGB}{187, 230, 240}    
\definecolor{myYellow}{RGB}{255, 242, 204}    
\definecolor{myOrange}{RGB}{255, 229, 204}    
\definecolor{myPurple}{RGB}{224, 208, 250}    

\definecolor{GrayLight}{gray}{0.35}

\newcommand{\cmark}{\ding{51}}%
\newcommand{\xmark}{\ding{55}}%

\newcommand{\ours}{JARVIS\xspace}

\def \ie {\emph{i.e.}}
\def \eg {\emph{e.g.}}

\newcommand{\tit}[1]{\smallbreak\noindent\textbf{#1.}}
\newcommand{\tinytit}[1]{\noindent\textbf{#1.}}

\newcommand{\llm}{\mathcal{G}}
\newcommand{\visenc}{\mathcal{F}_v}
\newcommand{\ctxvisenc}{\mathcal{F}_v^{\text{ctx}}}
\newcommand{\tgtvisenc}{\mathcal{F}_v^{\text{tgt}}}
\newcommand{\proj}{\mathrm{proj}}
\newcommand{\tgtproj}{\mathrm{proj}^{\text{tgt}}}
\newcommand{\visembd}{\mathbf{v}}
\newcommand{\visembdtgt}{\mathbf{\tilde{v}}}
\newcommand{\imglen}{\mathrm{N}}

\newcommand{\inputimg}{\text{I}}
\newcommand{\inputtxt}{\mathbf{x}}
\newcommand{\lossntp}{\mathcal{L}_{\text{NTP}}}
\newcommand{\lossjepa}{\mathcal{L}_{\text{JEPA}}}
\newcommand{\lossntpjepa}{\tilde{\mathcal{L}}_{\text{NTP}}}
\newcommand{\ctxblock}{\mathcal{M}^{\text{ctx}}}
\newcommand{\imgctx}{I^{\text{ctx}}}
\newcommand{\tgtblock}{\mathcal{M}^{\text{tgt}}}
\newcommand{\imgtgt}{I^{\text{tgt}}}
\newcommand{\latenttgt}{\mathbf{z}^{\text{tgt}}}
\newcommand{\latent}{\mathbf{z}}
\newcommand{\attntgttotgt}{\mathrm{attn}_{\text{tgt}\rightarrow\text{tgt}}}
\newcommand{\nattntgttotgt}{\mathrm{attn}_{\text{tgt}\nrightarrow\text{tgt}}}

\newcommand{\nattntgttotxt}{\mathrm{attn}_{\text{tgt}\nrightarrow\text{txt}}}
\newcommand{\attnours}{\mathrm{attn}_{\text{tgt}\nrightarrow\text{tgt},\text{tgt}\rightarrow\text{txt}}}

\definecolor{ourrow}{RGB}{224,235,255}     
\definecolor{groupband}{RGB}{235,242,250}  

\newcommand{\inc}[1]{\textcolor{blue!70!black}{\textbf{\footnotesize$\Delta$+#1}}}

\definecolor{cvprblue}{rgb}{0.21,0.49,0.74}
\definecolor{prova}{HTML}{5C8EC9}
\usepackage[pagebackref,breaklinks,colorlinks,allcolors=cvprblue]{hyperref}

\def\confName{CVPR}
\def\confYear{2026}

\title{Seeing Beyond Words: Self-Supervised Visual Learning for\\Multimodal Large Language Models}

\author{Davide Caffagni$^1$ \quad Sara Sarto$^1$ \quad Marcella Cornia$^1$ \quad Lorenzo Baraldi$^1$ \\Pier Luigi Dovesi$^2$ \quad Shaghayegh Roohi$^2$ \quad Mark Granroth-Wilding$^2$ \quad Rita  Cucchiara$^{1}$  \\
$^1$University of Modena and Reggio Emilia \quad $^2$AMD Silo AI\\
{\tt\small $^1$\{name.surname\}@unimore.it \quad $^2$\{name.surname\}@amd.com}
}

\begin{document}
\maketitle
\begin{abstract}
Multimodal Large Language Models (MLLMs) have recently demonstrated impressive capabilities in connecting vision and language, yet their proficiency in fundamental visual reasoning tasks remains limited. This limitation can be attributed to the fact that MLLMs learn visual understanding primarily from textual descriptions, which constitute a subjective and inherently incomplete supervisory signal. Furthermore, the modest scale of multimodal instruction tuning compared to massive text-only pre-training leads MLLMs to overfit language priors while overlooking visual details. To address these issues, we introduce \ours, a JEPA-inspired framework for self-supervised visual enhancement in MLLMs. Specifically, we integrate the I-JEPA learning paradigm into the standard vision-language alignment pipeline of MLLMs training. Our approach leverages frozen vision foundation models as context and target encoders, while training the predictor, implemented as the early layers of an LLM, to learn structural and semantic regularities from images without relying exclusively on language supervision. Extensive experiments on standard MLLM benchmarks show that \ours consistently improves performance on vision-centric benchmarks across different LLM families, without degrading multimodal reasoning abilities. Our source code is publicly available at: 
{\small{\url{https://github.com/aimagelab/JARVIS}}}.
\end{abstract}
    
\section{Introduction}
\label{sec:intro}

\begin{figure}[t]
\vspace{-.1cm}
    \centering
    \includegraphics[width=\linewidth]{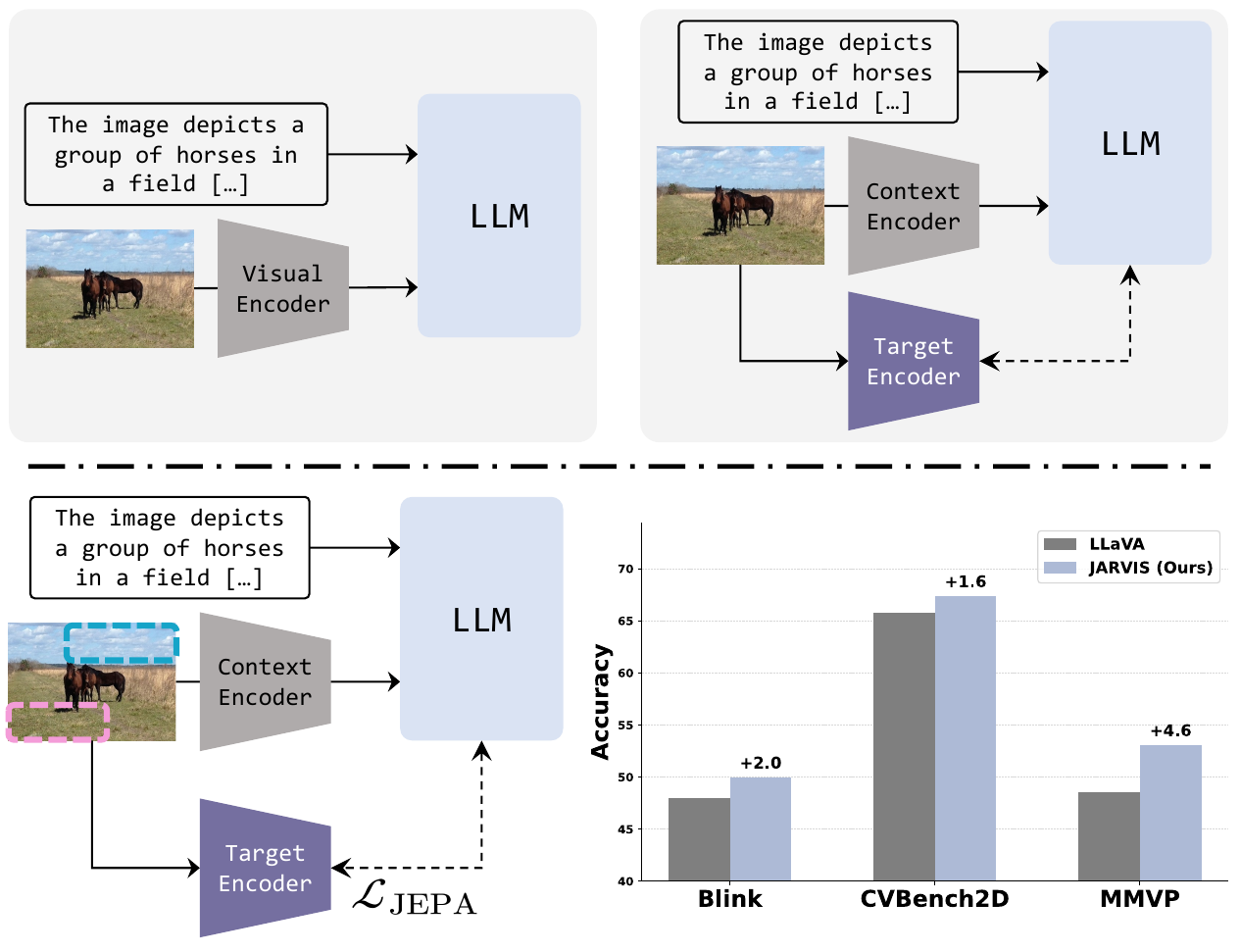}
    \vspace{-0.6cm}
    \caption{Comparison of LLaVA (top-left), a baseline that aligns the output of a selected layer with the output of a target encoder (top-right), and \ours that align the outputs employing a masked predictive loss (bottom-left). We also report the results of \ours and LLaVA across three vision benchmarks (bottom-right).}
    \label{fig:first}
    \vspace{-0.35cm}
\end{figure} 

The rapid success of Large Language Models (LLMs)~\cite{brown2020language,touvron2023llama,team2024qwen2} has shown the growing need for these models to process and reason across modalities beyond text. This demand has led to the emergence of Multimodal Large Language Models (MLLMs)~\cite{caffagni2024r}, which convert different input modalities into the same embedding space of the LLM, effectively allowing it to understand~\cite{alayrac2022flamingo, li2023blip, yemplug}, or even generate~\cite{team2024chameleon}, other modalities, with particular emphasis on images. Despite impressive progress, the fundamental recipe to design an MLLM has not changed since the introduction of visual instruction tuning, originally proposed by LLaVA~\cite{liu2023visual, liu2024improved, lin2024vila,cocchi2025llava}. LLaVA demonstrates that a lightweight projector can bridge visual and textual modalities by aligning the image representations from the vision encoder with the textual embedding space of the LLM.
Through this alignment, projected visual features can be effectively interpreted by the LLM, enabling it to reason about images and generate text conditioned on visual content. 

While this pipeline has proven highly effective across a broad range of tasks, current MLLMs still exhibit notable limitations in surprisingly simple visual reasoning scenarios, such as confirming the presence of objects, counting them, understanding their spatial relationships, or estimating their relative distance~\cite{grok, tong2024eyes, fu2024blink, tong2024cambrian}. The low proficiency of current MLLMs in these visual tasks highlights a severe deficit in their visual perception. We believe that this flaw emerges because MLLMs are trained to see images only via their textual descriptions. Indeed, during the alignment stage proposed by LLaVA, the MLLM is presented with an image, and the learning objective is to generate its caption. Intuitively, if the MLLM can describe an image, then it should have \textit{seen} it. However, image captions are inherently subjective~\cite{cheng2025caparena, sarto2025image}: they reflect what annotators
\textit{think} are relevant, often omitting details that may be crucial from other perspectives. Moreover, it is not practically feasible to assume having access to all possible descriptions of an image. 
Consequently, an image intrinsically contains richer and more comprehensive information than any subset of its textual descriptions. At the same time, because multimodal training is relatively modest compared to the massive unsupervised pre-training on textual corpora, MLLMs often over-rely on language priors when reasoning about an image, thereby overlooking visual details~\cite{zhang2025cross, wang2025towards, bi2025unveiling, deng2025words}. 

With that in mind, in this work we advocate for training MLLMs with the self-supervised signal inherently present within images, letting them discover and understand visual inputs beyond textual descriptions. In particular, building on the promising results on self-supervised learning achieved by Joint Embedding Predictive Architectures (JEPAs)~\cite{lecun2022path}, we propose to integrate I-JEPA~\cite{assran2023self}, a JEPA model trained on unlabeled images, into the classical visual instruction tuning pipeline. 

Specifically, given an image, we start by masking out contiguous blocks of its embeddings extracted by a pre-trained \textit{context} visual encoder. The remaining embeddings are forwarded to the LLM, which is trained to generate the image caption from the masked image, as usual with LLaVA. But concurrently, we also ask the LLM to recover the missing embeddings from the visible ones. Importantly, the supervision for these \textit{target} embeddings is provided by an additional target visual encoder, which only has access to the image, so that there is no language conditioning from its caption. This learning paradigm is backed by predictive coding theory~\cite{rao1999predictive}, which postulates that the representation of compatible parts of a signal (\ie, the visible context and the masked targets embeddings) should be predictable from each other. This design allows the LLM to learn about existing patterns and structures in images that often cannot be caught from their textual descriptions alone. An overview of our approach, that we name \ours, is given in Fig.~\ref{fig:first}.

In summary, this work highlights the importance and benefits of incorporating self-supervised visual learning into MLLMs. Given the same training data and architecture, our method, \ours, outperforms the original LLaVA training paradigm on standard MLLM benchmarks, with the most notable gains on the vision-centric tasks in the Cambrian evaluation suite~\cite{tong2024cambrian}. Our experiments show that the improvements are consistent across different families of LLMs and do not sacrifice general cognition performance.

\section{Related Work}
\label{sec:related}

\tinytit{Multimodal Large Language Models (MLLMs)} 
Multimodal Large Language Models (MLLMs)~\cite{alayrac2022flamingo, li2023blip, yemplug, team2024chameleon} have emerged as the natural extension of LLMs to modalities other than text. Depending on the extra modality of interest, most MLLMs feature a dedicated unimodal encoder and an adapter, whose role is to convert unimodal embeddings from the encoder into the LLM embedding space, which serves as the reasoning backbone~\cite{caffagni2024r}. Concerning image perception, the majority of models, including proprietary, state-of-the-art MLLMs~\cite{achiam2023gpt, team2023gemini, bai2025qwen2}, can be brought back to the two-stage training recipe proposed by LLaVA~\cite{liu2023visual, liu2024improved}, which builds upon an off-the-shelf LLM. During the first stage, the LLM is kept frozen, and only the image-to-text adapter (\eg, a linear or an MLP projector) is trained on image caption data to maximize the LLM probability of generating a caption given its image. Subsequently, the visual instruction tuning stage teaches the unfrozen LLM to follow users’ instructions in the presence of images. In this work, we adhere to the LLaVA framework, as it provides a training recipe that can be easily applied to multiple LLMs, based exclusively on open-source data. However, we intervene on the first training stage by adding a learning signal that does not involve language supervision.

\tit{Vision-centric Strategies for MLLMs}
Enhancing the visual perception of MLLMs is a long-standing challenge. Some methods focus on the vision encoder, designing image-to-text projectors that gather multi-layer activations before entering the LLM~\cite{chen2024lion, lin2025multi}. Another line of work integrates the CLIP vision encoder, which powers most MLLMs, with different vision experts for improved performance on vision domains~\cite{lin2023sphinx, kar2024brave, tong2024eyes, tong2024cambrian}. Concerning the language backbone, studies on how LLMs digest visual tokens have found that a limited number of attention heads are involved in processing visual tokens~\cite{bi2025unveiling}, and that they can be exploited to boost visual tasks such as grounding~\cite{kang2025your}. 

Self-supervised visual learning, in the form of image reconstruction, has also been explored, where an external vision encoder is only used as a teacher during training and discarded at inference time. ROSS~\cite{wangreconstructive} trains the MLLMs along with a denoiser network to recover the visual embeddings generated by a visual tokenizer. Closer to us is VIRAL~\cite{yoon2025visual}, which aligns the intermediate activations of the LLM with those from an external vision foundation model. 

In contrast to related efforts, we are the first, to the best of our knowledge, to integrate an I-JEPA-like supervision~\cite{assran2023self} as an additional learning objective in the training recipe of MLLMs. Specifically, besides aligning intermediate activations, we also ask the MLLM to recover the representation of masked portions of images, and notably, we apply it during the LLaVA pre-training stage, while the LLM is still learning how to process and understand visual tokens. This way, we influence how the LLM perceives images, going beyond pure language supervision.

\tit{Joint Embedding Predictive Architecture}
JEPA~\cite{lecun2022path} is an emerging framework for self-supervised learning, grounded on the principle that the latent representations of compatible input signals should be predictive of each other~\cite{rao1999predictive}. Depending on the domain of application~\cite{abdelfattah2024s, saito2025point, thimoniert, leim3, huang2025llm}, JEPA models differ in how to identify these pairs of compatible inputs. For instance, I-JEPA~\cite{assran2023self} trains context and target encoders so that the target representation of an image crop can be inferred from a predictor network, conditioned on the context representations of the nearby parts of the same image.
Later on, V-JEPA~\cite{bardesrevisiting, assran2025v} extends this idea to video, employing spatio-temporal blocks of masked patches to condition the predictor to output their unmasked representations from the target encoder. 

Typically, JEPA jointly learns both the context and the target networks. However, it has been recently shown that JEPA can learn strong representations efficiently even with a frozen target encoder~\cite{li2025rethinking}. In this work, we apply I-JEPA as a self-supervised objective to learn from images beyond their textual descriptions. While most JEPA-based methods focus on learning the encoder network, discarding the predictor network at inference, we leverage vision foundation models~\cite{radford2021learning, oquab2024dinov2} as frozen context and target encoders, and let an LLM be the predictor in our JEPA-augmented model.

\section{Proposed Method}
\label{sec:method}

\subsection{Background}
In this work, we aim to improve the visual perception capabilities of MLLMs. Besides a pre-trained LLM $\llm$, and without loss of generality, the architecture of an MLLM comprehends (i) a pre-trained visual encoder $\visenc$ to process images; and (ii) a trainable projector $\proj$, which aligns the output embedding space of $\visenc$ with the input embedding space of $\llm$. 
Formally, a given image $\inputimg$ is transformed into a sequence of $\imglen$ $d$-dimensional visual embeddings, which are then provided as input tokens to $\llm$, as follows:
\begin{align*}
    I &= \proj(\visenc(\inputimg)) \\
    & = \{\visembd_i \in \mathrm{R}^d\}_{i=1\dots\imglen}\}.
\end{align*}

\begin{figure}[t]
    \centering    \includegraphics[width=0.98\linewidth]{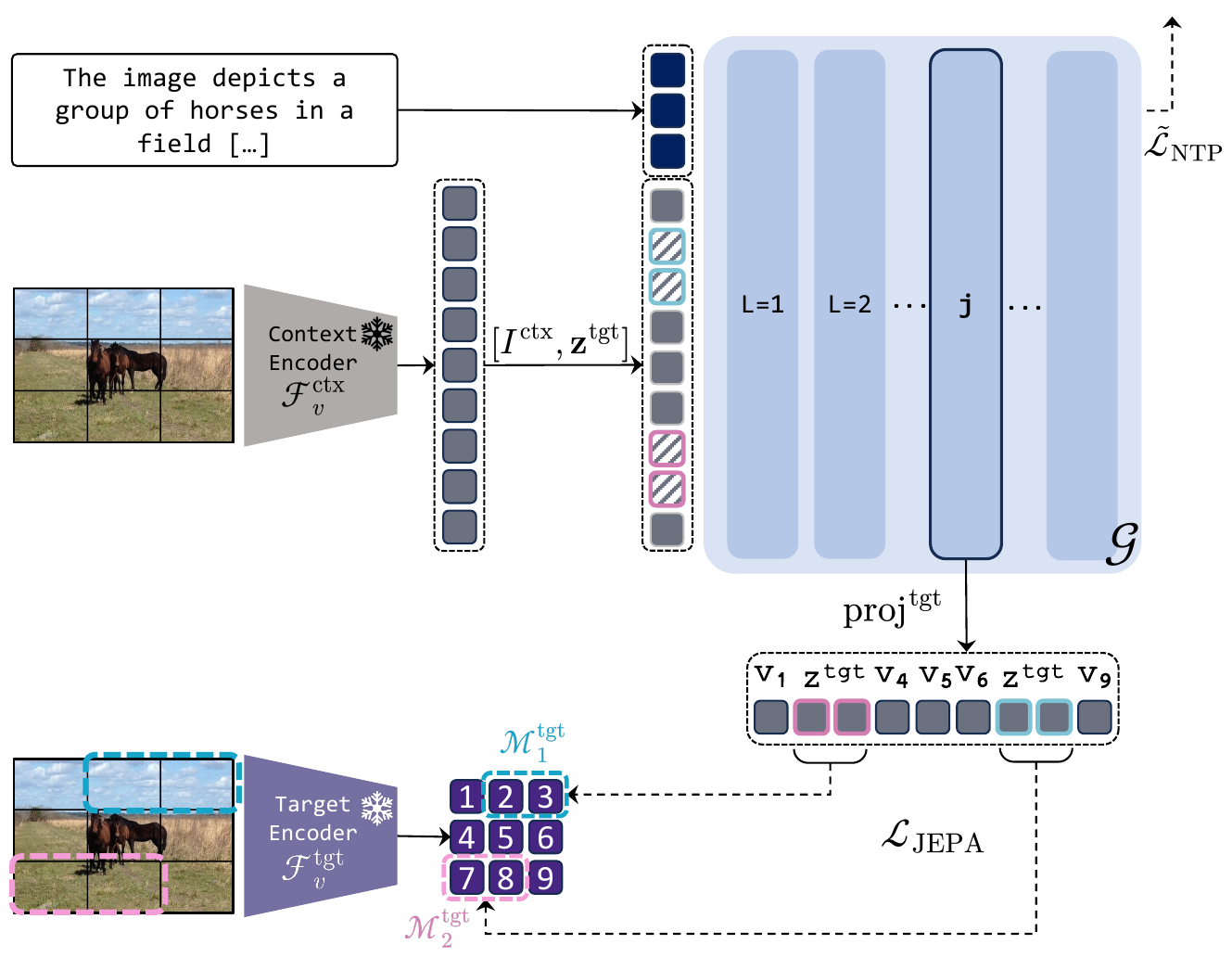}
    \vspace{-0.1cm}
    \caption{Overview of our \textbf{\ours} method. We leverage a single context block to predict the representations of multiple target blocks through a masked predictive loss, aligning the predicted embeddings of the LLM with the outputs of a target encoder.}
    \label{fig:model}
    \vspace{-0.2cm}
\end{figure}

Following the approach popularized by LLaVA~\cite{liu2023visual}, we train $\llm$ with a next-token prediction (NTP) objective, specifically by minimizing the negative log-likelihood of generating text $\inputtxt$ while being conditioned on image $\inputimg$:
\begin{equation}
    \label{eq:loss_ntp}
    \lossntp = - \sum_{i} \log P\left(\inputtxt_i | I, \inputtxt_{1,\dots,i-1} ; \llm, \proj\right).
\end{equation}

The training process is divided into two stages. During the first stage, referred to as \textit{alignment}, $\inputtxt$ represents the caption of $\inputimg$, and only $\proj$ is trained to align visual and textual representations. In the second stage, known as \textit{visual instruction tuning}, $\inputtxt$ represents a multi-turn visual dialog concerning the image $\inputimg$, during which both the language model $\llm$ and the projector $\proj$ are trained.

\subsection{Bringing JEPA into MLLMs}
By minimizing Eq.~\ref{eq:loss_ntp} during the alignment stage, the MLLM learns to describe images in natural language. Intuitively, succeeding on this task implies that the model has acquired the ability to \textit{see} and understand the image. However, textual supervision alone is inherently limited, as captions cannot fully capture the rich visual information embedded within an image. To address this limitation, we propose to integrate a purely visual self-supervised learning objective. Specifically, inspired by the recent progress of JEPA~\cite{lecun2022path} models, we integrate the I-JEPA~\cite{assran2023self} formulation into the alignment stage of LLaVA~\cite{liu2023visual,liu2024improved}.

\begin{figure*}[t]
  \label{fig:image1}
  \centering
  \begin{minipage}[t]{0.505\textwidth}
    \centering
        \includegraphics[width=\linewidth]{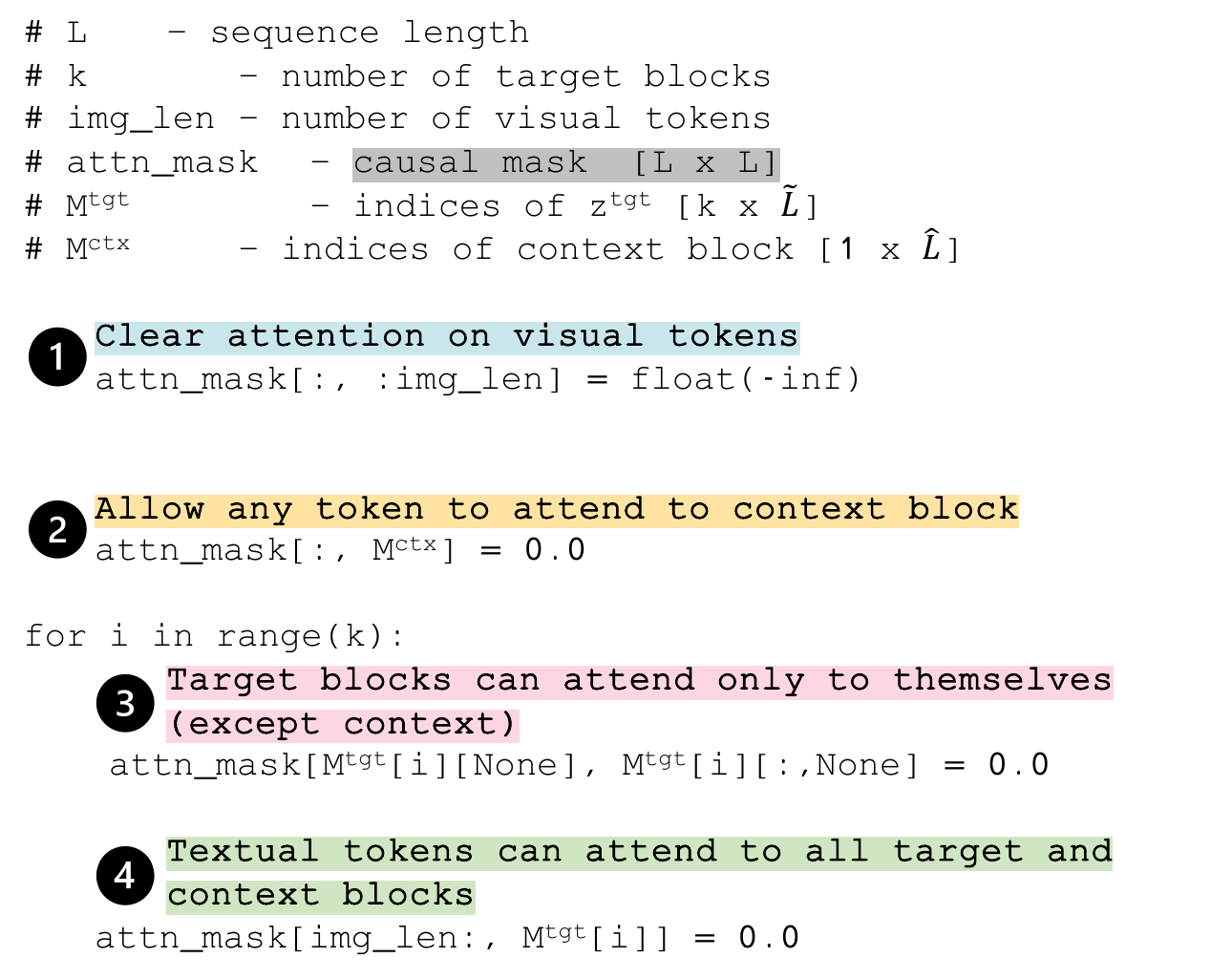}
  \end{minipage}%
  \hspace{0.005cm}
  \begin{minipage}[t]{0.485\textwidth}
    \centering
        \includegraphics[width=\linewidth]{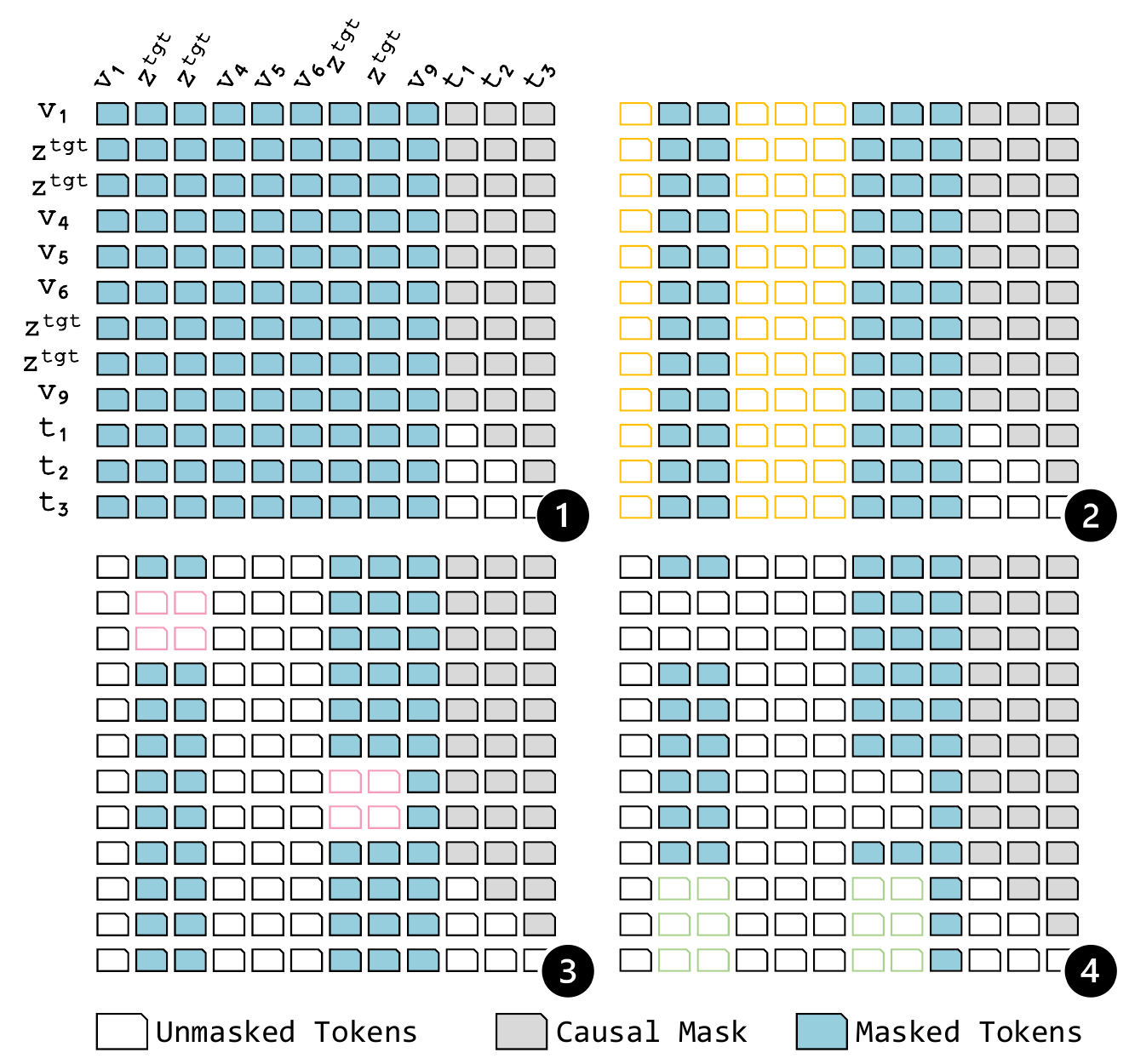}
        \end{minipage}
  \vspace{-0.1cm}
  \caption{Visualization of the attention mask implementation. \textbf{Left:} pseudo-code outlining the sequential steps used to compute the attention mask. \textbf{Right:} graphical illustration showing the effect of each step, highlighting how the mask modifies token attention.}
\label{fig:double_placeholder}
\vspace{-.2cm}
\end{figure*}

\tit{Revisiting the I-JEPA Approach}
The goal of I-JEPA is to learn a visual embedding function such that the latent representation of missing parts of an image, \ie, the targets, can be predicted from the representation of the visible part, \ie, the context. Given an image $\inputimg$, I-JEPA begins by sampling a set of integers $\ctxblock = \{ i \subseteq 1,\dots,\imglen \}$, each corresponding to the index of an image patch in $\inputimg$, that build up the visible context. Similarly, $k$ different sets of patch indices $\tgtblock = \{\tgtblock_j \subseteq 1,\dots,\imglen\}_{j=1,\dots,k}$ are drawn to serve as the prediction targets, and thus will be masked out from $\inputimg$. As it can be seen in Fig.~\ref{fig:model} (bottom left), these sets correspond to blocks of contiguous image patches, whose aspect ratio and size (and therefore the cardinality of $\ctxblock$ and $\tgtblock$) are sampled from a predefined range. Importantly, while target blocks can overlap with each other, there must not be intersection between context and targets. 

Unlike I-JEPA, we do not mask the input image at the pixel level, but directly at the visual embeddings extracted by the frozen visual encoder $\visenc$, which -- for clarity -- will be referred to as $\ctxvisenc$ in the following. Further, an additional, frozen visual encoder $\tgtvisenc$ is employed to extract targets embeddings $\imgtgt$ from the set of target blocks. Formally, the two sets of context and target visual embeddings are defined as follows:
\begin{align}
\label{eq:ctx_tgt_block}
\imgctx &= \{ \visembd_i \in \proj(\ctxvisenc(\text{I})) \text{ s.t. } i \in \ctxblock \} \\
\imgtgt &= \{ \visembdtgt_i \in \tgtvisenc(\text{I}) \text{ s.t. } i \in \tgtblock \} .
\end{align}

According to I-JEPA, a predictor network is trained to predict $\imgtgt$ from $\imgctx$ and a latent variable $\latenttgt$, which signal the position of the masked target patches. In practice, $\latenttgt$ is implemented as a learnable embedding $\latent \in \mathrm{R}^d$ summed with the positional encoding $\boldsymbol{\phi}(\cdot) \in \mathrm{R}^d$ of each masked patch:
\begin{equation}
    \label{eq:latents}
    \latenttgt = \{\latent + \boldsymbol{\phi}(i), \ \forall i \in \tgtblock\}.
\end{equation}

\tit{Layer-wise Target Prediction}
In our model, the predictor network corresponds to the LLM $\llm$, so that it can learn to recover the embeddings of the masked image patches, thus discovering latent patterns and structures of images. Specifically, as recent works~\cite{bi2025unveiling, zhang2025cross, wang2025towards} have demonstrated that MLLMs concentrate attention to visual tokens in their early-to-middle layers, we select a shallow subset from the stack of Transformer layers of $\llm$ to implement the predictor. Let $\llm_j(\cdot)_i$ be the activation from the chosen layer $j$ at the $i$-th token position. We can now define the JEPA loss function to be minimized as a distance measure between the predictor output at the target positions and the reference targets $\imgtgt$. Formally, this can be expressed as:
\begin{equation}
    \label{eq:loss_jepa}
    \lossjepa = \frac{1}{|\tgtblock|} \sum_{i \in \tgtblock} d\left(\tgtproj(\llm_j([\imgctx, \latenttgt]))_i, \imgtgt_i\right),
\end{equation}
where $d(\cdot)$ is a distance function, $\tgtproj$ is a feed-forward network that matches the dimension of $\llm_j(\cdot)$ with that from $\tgtvisenc$, and $[\cdot]$ is the concatenation over the token axis.

At the same time, we still train our MLLM to generate the image caption $\inputtxt$, while having access to the visible context $\imgctx$ and the latents $\latenttgt$. Notably, thanks to $\lossjepa$, layer after layer, the representation of $\latenttgt$ collects information about the missing part of the image, easing the challenge of captioning a masked image. The next-token prediction objective of Eq.~\ref{eq:loss_ntp} is thus modified as follows:
\begin{equation}
    \small
    \label{eq:loss_ntp_jepa}
    \lossntpjepa = - \sum_{i} \log P\left(\inputtxt_i | [\imgctx, \latenttgt], \inputtxt_{1,\dots,i-1} ; \llm, \proj\right).
\end{equation}

\tit{Efficient Implementation via Attention Mask} The overall training objective is given by the sum of $\lossntpjepa$ and $\lossjepa$. Note that the two losses are computed in a single forward pass, without requiring any additional compute on the LLM $\llm$.  This is achieved by tweaking the attention mask applied within each attentive layer of $\llm$ as follows (depicted in Fig.~\ref{fig:double_placeholder}). First, because context and target visual embeddings can be intertwined within the visual input sequence $[\imgctx, \latenttgt]$, we allow bidirectional attention among them, with the caveat that (i) context embeddings $\imgctx$ cannot attend to targets $\imgtgt$, according to I-JEPA; and (ii) target embeddings from different blocks cannot attend to each other, while being free to attend to $\imgctx$. Next, because the image caption tokens $\inputtxt$ obey the causal attention mechanism of $\llm$, and they follow the visual embeddings in the input sequence, the visual context $\imgctx$ and the latents $\latenttgt$ are not influenced by the image caption $\inputtxt$, so there is no textual supervision while predicting $\imgtgt$.

\tit{Balancing Visual and Textual Supervision} Given the same number of optimization steps, training an MLLM on exclusively masked images would produce a weaker image-text alignment. Moreover, this would create an inconsistency between the alignment stage and the subsequent visual instruction tuning and inference stages, where the model has access to whole images. To account for that, we propose to skip the computation of $\lossjepa$ with probability $\lambda$, thus computing the original next-token prediction loss (cf. Eq.~\ref{eq:loss_ntp}) conditioned on the unmasked image. This is equivalent to extending $\imgctx$ to the entire image, thereby removing the need to insert the latent variable $\latenttgt$, since there are no masked patches to reconstruct.

\smallskip
\noindent{Following LLaVA, once the alignment training is completed, we proceed to the visual instruction tuning stage. In this setting, the target visual encoder $\tgtvisenc$ and projector $\tgtproj$ are not used anymore, and the model is trained with the next-token prediction objective on unmasked images.}

\section{Experiments}

\tinytit{Implementation and Training Details} We train \ours, as well as the baselines and other methods, by strictly following the LLaVA-1.5~\cite{liu2024improved} recipe, and only applying the changes required to implement the proposed approach. When not specified otherwise, all models feature CLIP ViT-L/14@336~\cite{radford2021learning} as the (context) visual encoder $\ctxvisenc$, while the image-text projector $\proj$ is implemented as a two-layer MLP network. In the alignment stage, which comprises 558k image-caption pairs, we leave the LLM $\llm$ frozen and only train $\proj$, along with the target projector $\tgtproj$ for \ours. Conversely, $\llm$ is unfrozen while training for visual instruction tuning on the 665k samples from the dataset released with LLaVA-1.5. Differently, the visual encoders are always kept frozen in both stages. Further training hyperparameters are kept consistent across all models and are detailed in the supplementary material.

To select the context $\ctxblock$ and target $\tgtblock$ (cf. Eq~\ref{eq:ctx_tgt_block}) indices of visual embeddings, we implement the exact block-wise masking strategy proposed by I-JEPA~\cite{assran2023self}. For $\ctxblock$, we sample patch indices to get a single block of  visual embeddings, corresponding to a rectangular crop of the input image, covering from 85\% up to 100\% of the picture. On the other hand, we sample $k=4$ sets of indices, corresponding to crops of a smaller size than $\ctxblock$, and whose union builds up the indices of the target embeddings $\tgtblock$, that are removed from $\ctxblock$ to prevent trivial predictions.

To implement the predictor network and compute $\lossjepa$, we leverage the first quarter of the layers from $\llm$, \ie~we set $j$ (cf. Eq.~\ref{eq:loss_jepa}) to the index of the layer at one fourth of the depth for each specific LLM. The architecture of $\tgtproj$ follows the one from $\proj$~\cite{liu2024improved}, that is, a two-layer MLP with a GELU non-linearity~\cite{hendrycks2016gaussian} inside. Concerning the target visual encoder $\tgtvisenc$, if not specified differently, we use DINOv2-L/14~\cite{oquab2024dinov2}, known to deliver high-quality spatial features. During alignment training, we set $\lambda$ equal to $0.2$ in all experiments, meaning that we skip the optimization step for $\lossjepa$ 20\% of the times. Finally, we use the negative cosine similarity as the distance function $d(\cdot;\cdot)$ between predicted and target visual embeddings in Eq.~\ref{eq:loss_jepa}.

\begin{table*}[t]
\centering
\caption{Ablation study results. All experiments employ Vicuna-7B as the underlying LLM with CLIP ViT-L@336 as context visual encoder and DINOv2-L/14 as target visual encoder.}
\vspace{-0.15cm}
\setlength{\tabcolsep}{.35em}
\resizebox{\linewidth}{!}{%
\begin{tabular}{lc c c c c c c >{\columncolor{ourrow!30}}c>{\columncolor{ourrow!30}}c>{\columncolor{ourrow!30}}c>{\columncolor{ourrow!30}}c>{\columncolor{ourrow!30}}c>{\columncolor{ourrow!30}}c>{\columncolor{ourrow!30}}c}
\toprule 
& & \textbf{General} & & \textbf{Knowledge} & & \textbf{OCR} & & \multicolumn{6}{c}{\cellcolor{ourrow!30}\textbf{Vision-Centric}} \\
\cmidrule(lr){3-3} \cmidrule(lr){5-5} \cmidrule(lr){7-7} \cmidrule{9-14}
& & Avg & &  Avg & & Avg && RealWorldQA & MMVP & Blink & CVBench2D & CVBench3D & Avg \\
\midrule 
LLaVA~\cite{liu2024improved} & & 65.8 & & 41.4 & & 36.2 & & 54.9 & 31.3 & 46.8 & 57.6 & 63.7 & 50.9 \\
\midrule
\rowcolor{gray!10}
\multicolumn{14}{l}{\textit{Effect of changing the LLM layer}} \\
$j=4$  (early) & & \textbf{65.1} & & 41.1 & & 36.1 && 55.9 & 28.7 & 45.8 & 57.7 & 64.1 & 50.4 \\
\rowcolor{ourrow}
$j=8$  (¼-depth) & & \textbf{65.1} & & 41.6 & & \textbf{36.2} & & 55.6 & \textbf{30.7} & \textbf{48.3} & 60.4 & \textbf{63.8} & \textbf{51.7} \\
$j=16$ (middle) & & 64.5 & & 41.7 & & \textbf{36.2} & & 54.6 & 28.0 & 45.3 & \textbf{60.8} & 63.0 & 50.3 \\
$j=24$ (¾-depth) & & 65.0 & & \textbf{42.3} & & \textbf{36.2} & & \textbf{57.0} & 26.7 & 46.5 & 59.4 & 61.5 & 50.2 \\
$j=31$ (near-final) & & 64.9 & & 41.9 & & 35.8 & & 55.7 & 27.3 & 45.5 & 56.6 & 60.9 & 49.2 \\
$j=32$ (final) & & 64.4 & & 41.7 & & 35.9 & & \textbf{57.0} & 27.3 & 45.4 & 55.5 & 61.7 & 49.4 \\
\midrule
\rowcolor{gray!10}
\multicolumn{14}{l}{\textit{Effect of projector scaling}} \\
w/ linear projection & & 64.8 & & \textbf{42.1} & & 35.9 & & 54.9 & 26.0 & 45.3 & 58.3 & 63.7 & 49.6 \\
\rowcolor{ourrow}
w/ MLP projection & & \textbf{65.1} & & 41.6 & & \textbf{36.2} & & \textbf{55.6} & \textbf{30.7} & \textbf{48.3} & \textbf{60.4} & \textbf{63.8} & \textbf{51.7} \\
\midrule
\rowcolor{gray!10}
\multicolumn{14}{l}{\textit{Effect of changing the distance function in $\lossjepa$}} \\
w/ smooth L1 distance (I-JEPA~\cite{assran2023self}) & & 64.7 & & \textbf{41.9} & & 35.8 & & \textbf{56.2} & 24.7 & 45.5 & 58.1 & 61.1 & 49.1 \\
\rowcolor{ourrow}
w/ cosine distance & & \textbf{65.1} & & 41.6 & & \textbf{36.2 }& & 55.6 & \textbf{30.7} & \textbf{48.3} & \textbf{60.4} & \textbf{63.8} & \textbf{51.7} \\
\bottomrule
\end{tabular}
}
\label{tab:ablation}
\vspace{-0.3cm}
\end{table*}

\tit{Evaluation Benchmarks}
For evaluation, we employ the Cambrian evaluation suite~\cite{tong2024cambrian}, a comprehensive benchmark comprising 16 tasks spanning four categories: General (4), Knowledge (4), OCR (3), and Vision-Centric (5). Concerning the Vision-Centric category, Cambrian comprehends RealWorldQA~\cite{grok}, which evaluates commonsense reasoning based on visual inputs; MMVP~\cite{tong2024eyes}, which probes the visual perception skills of a model across nine classes of questions; Blink~\cite{fu2024blink} and CVBench2D~\cite{lin2014microsoft,zhou2019semantic}, which focus on questions concerning spatial relationships; and CVBench3D~\cite{brazil2023omni3d}, which evaluates the ability of a model to assess the relative depth of objects from the camera, as well as the relative distance between objects. For each task, we report the accuracy computed with the official evaluation toolkit. Details on the General, Knowledge, and OCR categories are provided in the supplementary, along with a breakdown of the experimental results on each task.

\subsection{Ablation Studies}
We start by presenting a set of ablation studies designed to understand how key architectural and objective choices influence the behavior of \ours. For these experiments, we employ Vicuna-7B as the underlying LLM. 

\tit{Varying the LLM Layer for $\lossjepa$} When LLMs are applied to language tasks, activations from their final layer are typically extracted for token generation. However, when computing $\lossjepa$ (cf. Eq.~\ref{eq:loss_jepa}), \ours also leverages an LLM to predict the missing part of an image in a latent space. We investigate whether this learning objective is better-suited for computation on intermediate layers, given that the last layer should be more sensitive to the syntax and grammar structures crucial for producing coherent linguistic output. Table~\ref{tab:ablation} (top) reports the performance while varying the layer at which $\lossjepa$ is computed. Note that $\lossntpjepa$ (cf. Eq.~\ref{eq:loss_ntp_jepa}) is always computed on the final layer. As a \textit{baseline}, we also include in the first row the evaluation of Vicuna-7B when trained according to LLaVA~\cite{liu2024improved}.

While the average accuracy on General, Knowledge, and OCR tasks remains stable, there is a striking difference on visual tasks when using intermediate instead of final layer activations. For instance, Vision-Centric accuracy increases by $+1.0$ point when descending from layer $31$ to layer $24$. Among the intermediate layers, we found the one located at one fourth of the depth (\ie, $j=8$ for Vicuna-7B, which is $32$ layers depth) to deliver the best visual performance, with strong improvements on MMVP and CVBench2D of $+3.4$ and $+4.9$ points respectively, against the final layer. 
These findings support our earlier observations and align with recent studies~\cite{bi2025unveiling, wang2025towards, zhang2025cross}, which show that MLLMs focus more on visual tokens in intermediate layers.

\tit{Scaling the Target Projector}
Next, we ablate the impact of scaling the target projector $\tgtproj$ from a simple linear layer to a two-layer MLP in Table~\ref{tab:ablation} (middle). $\tgtproj$ serves to convert the intermediate activation of the LLM out of the $j$-th layer, \ie, $\llm_j(\cdot)$, so as to match the same dimensionality of the target embeddings $\imgtgt$ generated by $\tgtvisenc$ for computing $\lossjepa$. Using a linear projection not only fails to improve visual performance, but even degrades it compared to the baseline. For instance, on the Vision-Centric Blink benchmark, performance drops from $46.8$ to $45.3$. In contrast, employing an MLP projector improves performance, yielding a $+1.7$ points gain on Vision-Centric tasks. This finding suggests that a linear transformation is not enough to project intermediate LLM embeddings into the output space of a visual encoder.

\tit{Changing the Distance Function in $\lossjepa$}
Optimizing for $\lossjepa$ equals minimizing the distance between the predicted and target visual embeddings. By default, we choose the negative cosine similarity (\ie, the cosine distance) as our distance function. It follows that we are asking the LLM to guess the direction of the target embeddings. Conversely, we explore the effect of switching to the smooth L1 distance measure, as done in the original I-JEPA formulation~\cite{assran2023self}, which not only requires matching the direction of the prediction targets, but also their magnitude, in order to be minimized. Specifically, the smooth L1 distance decreases quadratically if the (element-wise) absolute distance between predicted and target embeddings is below $1$, while otherwise increasing linearly.
According to Table~\ref{tab:ablation} (bottom), switching from the cosine distance to the smooth L1 distance leads to a $+2.6$ average accuracy points on visual tasks. Critically, minimizing $\lossjepa$ with the smooth L1 distance severely degrades the General and Vision-Centric performance with respect to the baseline.

\smallskip
\noindent In the remaining of this work, we present experimental results that build on the insights from our ablations. To sum up, we leverage the first quarter of the LLM layers to implement the predictor network required for I-JEPA, extended with a two-layer MLP projector to match the dimensionality between the predicted and target visual embeddings, and measuring their distance with the negative cosine similarity.

\begin{table*}[t]
\centering
\caption{Comparison against competing approaches, including the original LLaVA~\cite{liu2024improved} model, VIRAL~\cite{yoon2025visual}, and our baselines.}
\vspace{-0.15cm}
\setlength{\tabcolsep}{.35em}
\resizebox{\linewidth}{!}{%
\begin{tabular}{lccc c c c c c >{\columncolor{ourrow!30}}c>{\columncolor{ourrow!30}}c>{\columncolor{ourrow!30}}c>{\columncolor{ourrow!30}}c>{\columncolor{ourrow!30}}c>{\columncolor{ourrow!30}}c>{\columncolor{ourrow!30}}c}
\toprule 
& & & \textbf{General} & & \textbf{Knowledge} & & \textbf{OCR} & & \multicolumn{6}{c}{\cellcolor{ourrow!30}\textbf{Vision-Centric}} \\
\cmidrule(lr){4-4} \cmidrule(lr){6-6} \cmidrule(lr){8-8} \cmidrule{10-15}
& \textbf{LLM} & & Avg & &  Avg & & Avg && RealWorldQA & MMVP & Blink & CVBench2D & CVBench3D & Avg\\
\midrule
LLaVA~\cite{liu2024improved} & Vicuna-7B & & \textbf{65.8} & & 41.4 & & 36.2 & & 54.9 & \textbf{31.3} & 46.8 & 57.6 & 63.7 & 50.9 \\
VIRAL~\cite{yoon2025visual} & & & 64.6 & & \textbf{42.6} & & 36.0 & & 54.8 & 30.0 & 42.9 & 55.7 & 60.0 & 48.7 \\
Ours w/o Masking & & & 65.0 & & 41.3 & & \textbf{36.3} & & \textbf{55.7} & 28.0 & 42.7 & 57.7 & 60.8 & 49.0 \\
\rowcolor{ourrow}
\textbf{\ours (Ours)} & & & 65.1 & & 41.6 & & 36.2 & & 55.6 & 30.7 & \textbf{48.3} & \textbf{60.4} & \textbf{63.8} & \textbf{51.7} \\
\midrule
LLaVA~\cite{liu2024improved} & Qwen2-7B & & 72.0 & & 46.7 & & 36.9 & & \textbf{58.0} & 36.7 & \textbf{52.7} & 62.5 & 66.8 & 55.3 \\
VIRAL~\cite{yoon2025visual} & & & 70.9 & & 47.2 & & 36.1 & & 57.9 & 36.0 & 49.3 & 62.0 & 62.3 & 53.5 \\
Ours w/o Masking & & & \textbf{72.4} & & \textbf{48.1} & & \textbf{37.0} & & 56.2 & 35.3 & 49.6 & 62.8 & 67.6 & 54.3 \\
\rowcolor{ourrow}
\textbf{\ours (Ours)} & & & 71.9 & & 47.7 & & 36.3 & & 55.8 & \textbf{38.0} & 49.3 & \textbf{63.9} & \textbf{73.0} & \textbf{56.0} \\
\bottomrule
\end{tabular}
}
\label{tab:competitors}
\vspace{-0.1cm}
\end{table*}

\begin{figure*}[t!]
\begin{minipage}{0.24\linewidth}
\scriptsize{\textbf{Q}: How many flowerpots are in the image?\\
\vspace{0.05cm}}\\
\begin{minipage}{0.443\linewidth}
\includegraphics[width=1.\linewidth,height=0.88\linewidth]{images/qualitatives/count_147.png}
\end{minipage}
\hfill
\begin{minipage}{0.53\linewidth}
\scriptsize{
\textbf{LLaVA~\cite{liu2024improved}}:\\
D) 3 \textcolor{red}{\xmark} \\
\textbf{VIRAL~\cite{yoon2025visual}}:\\
D) 3 \textcolor{red}{\xmark} \\
\textbf{\ours (Ours):}\\
E) 2 \textcolor[HTML]{00b050}{\cmark}
}
\end{minipage}
\end{minipage}
\hspace{0.01cm}
\begin{minipage}{0.24\linewidth}
\scriptsize{\textbf{Q}: Where is the bowl (red box) located\\ with respect to the teddy bear?
\vspace{0.05cm}}\\
\begin{minipage}{0.443\linewidth}
\includegraphics[width=1.\linewidth,height=0.88\linewidth]{images/qualitatives/relative_458.png}
\end{minipage}
\hfill
\begin{minipage}{0.53\linewidth}
\scriptsize{
\textbf{LLaVA~\cite{liu2024improved}}:\\
A) Above \textcolor{red}{\xmark} \\
\textbf{VIRAL~\cite{yoon2025visual}}:\\
A) Above \textcolor{red}{\xmark} \\
\textbf{\ours (Ours):}\\
B) Below \textcolor[HTML]{00b050}{\cmark}
}
\end{minipage}
\end{minipage}
\hspace{0.01cm}
\begin{minipage}{0.24\linewidth}
\scriptsize{\textbf{Q}: Which object is closer to the camera taking this photo, the lamp (red box) or the books (blue box)?
\vspace{0.05cm}}\\
\begin{minipage}{0.443\linewidth}
\includegraphics[width=1.\linewidth,height=0.88\linewidth]{images/qualitatives/depth_155.png}
\end{minipage}
\hfill
\begin{minipage}{0.53\linewidth}
\scriptsize{
\textbf{LLaVA~\cite{liu2024improved}}:\\
A) Lamp \textcolor{red}{\xmark} \\
\textbf{VIRAL~\cite{yoon2025visual}}:\\
A) Lamp \textcolor{red}{\xmark} \\
\textbf{\ours (Ours):}\\
B) Books \textcolor[HTML]{00b050}{\cmark}
}
\end{minipage}
\end{minipage}
\hspace{0.01cm}
\begin{minipage}{0.24\linewidth}
\scriptsize{\textbf{Q}: Given the two bounding boxes on the image, which one more accurately localizes and encloses the cat?
\vspace{0.05cm}}\\
\begin{minipage}{0.443\linewidth}
\includegraphics[width=1.\linewidth,height=0.88\linewidth]{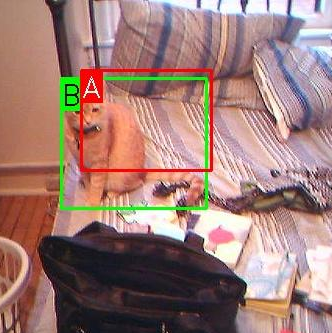}
\end{minipage}
\hfill
\begin{minipage}{0.53\linewidth}
\scriptsize{
\textbf{LLaVA~\cite{liu2024improved}}:\\
A) Box A \textcolor{red}{\xmark} \\
\textbf{VIRAL~\cite{yoon2025visual}}:\\
A) Box A \textcolor{red}{\xmark} \\
\textbf{\ours (Ours):}\\
B) Box B \textcolor[HTML]{00b050}{\cmark}
}
\end{minipage}
\end{minipage}
\vspace{0.2cm}

\begin{minipage}{0.24\linewidth}
\scriptsize{\textbf{Q}: How many skys are in the image?\\
\vspace{0.05cm}}\\
\begin{minipage}{0.443\linewidth}
\includegraphics[width=1.\linewidth,height=0.88\linewidth]{images/qualitatives/count_376.png}
\end{minipage}
\hfill
\begin{minipage}{0.53\linewidth}
\scriptsize{
\textbf{LLaVA~\cite{liu2024improved}}:\\
B) 0 \textcolor{red}{\xmark} \\
\textbf{VIRAL~\cite{yoon2025visual}}:\\
D) 2 \textcolor{red}{\xmark} \\
\textbf{\ours (Ours):}\\
C) 1 \textcolor[HTML]{00b050}{\cmark}
}
\end{minipage}
\end{minipage}
\hspace{0.01cm}
\begin{minipage}{0.24\linewidth}
\scriptsize{\textbf{Q}: Where is the dining table (red box) located with respect to the person?
\vspace{0.05cm}}\\
\begin{minipage}{0.443\linewidth}
\includegraphics[width=1.\linewidth,height=0.88\linewidth]{images/qualitatives/relative_309.png}
\end{minipage}
\hfill
\begin{minipage}{0.53\linewidth}
\scriptsize{
\textbf{LLaVA~\cite{liu2024improved}}:\\
A) Right \textcolor{red}{\xmark} \\
\textbf{VIRAL~\cite{yoon2025visual}}:\\
A) Right \textcolor{red}{\xmark} \\
\textbf{\ours (Ours):}\\
A) Left \textcolor[HTML]{00b050}{\cmark}
}
\end{minipage}
\end{minipage}
\hspace{0.01cm}
\begin{minipage}{0.24\linewidth}
\scriptsize{\textbf{Q}: Which  is closer to the table (red box),\\books (blue box) or shelves (green box)?
\vspace{0.05cm}}\\
\begin{minipage}{0.443\linewidth}
\includegraphics[width=1.\linewidth,height=0.88\linewidth]{images/qualitatives/distance_637.png}
\end{minipage}
\hfill
\begin{minipage}{0.53\linewidth}
\scriptsize{
\textbf{LLaVA~\cite{liu2024improved}}:\\
B) Shelves \textcolor{red}{\xmark} \\
\textbf{VIRAL~\cite{yoon2025visual}}:\\
B) Shelves \textcolor{red}{\xmark} \\
\textbf{\ours (Ours):}\\
A) Books \textcolor[HTML]{00b050}{\cmark}
}
\end{minipage}
\end{minipage}
\hspace{0.01cm}
\begin{minipage}{0.24\linewidth}
\scriptsize{\textbf{Q}: How many drive wheels are on the engine?
\vspace{0.05cm}}\\
\begin{minipage}{0.443\linewidth}
\includegraphics[width=1.\linewidth,height=0.88\linewidth]{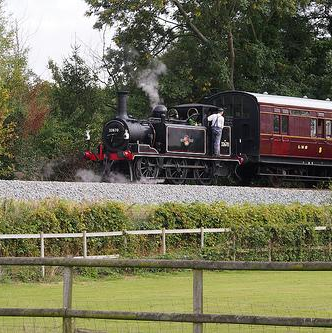}
\end{minipage}
\hfill
\begin{minipage}{0.53\linewidth}
\scriptsize{
\textbf{LLaVA~\cite{liu2024improved}}:\\
C) 2 \textcolor{red}{\xmark} \\
\textbf{VIRAL~\cite{yoon2025visual}}:\\
C) 2 \textcolor{red}{\xmark} \\
\textbf{\ours (Ours):}\\
D) 3 \textcolor[HTML]{00b050}{\cmark}
}
\end{minipage}
\end{minipage}
\vspace{-0.1cm}
\caption{Qualitative comparison of three training methods for MLLMs with Qwen2-7B~\cite{team2024qwen2}. We show samples from CVBench2D~\cite{tong2024cambrian} (Count and Relative, 1st and 2nd columns), CVBench3D~\cite{tong2024cambrian} (Depth and Distance, 3rd column), and Blink~\cite{fu2024blink} (last column).}
\label{fig:qualitatives}
\vspace{-.1cm}
\end{figure*}

\subsection{Main Experimental Results}

\tinytit{Comparison with Related Methods and Baselines}
We analyze different training recipes for MLLMs applied to two LLMs, Vicuna-7B~\cite{vicuna2023} (top), and Qwen2-7B~\cite{team2024qwen2} (bottom), collecting the results in Table~\ref{tab:competitors}. A part from the standard LLaVA~\cite{liu2024improved} (first row), all the other methods have been implemented with DINOv2-L/14~\cite{oquab2024dinov2} as target visual encoder. In particular, we add VIRAL~\cite{yoon2025visual} (second row), a novel method for visual representation alignment between LLMs and foundation visual encoders. VIRAL regularizes the visual instruction tuning stage of LLaVA with an additional objective to align the activation from the middle layer of the LLM with the output of an external target visual encoder. 

Additionally, we implement another baseline (third row), which follows the exact same architecture as \ours, but does not apply the masked predictive objective of I-JEPA~\cite{assran2023self}. Instead, we train it to simply align the activation from the layer at one-fourth of the depth of the LLM with the target visual encoder. While this is conceptually similar to VIRAL, the extra visual objective is applied during the alignment stage rather than visual instruction tuning, as in \ours. This baseline allows us to measure the impact of $\lossjepa$ on our training method. In other words, that comparison answers the question: to unlock the visual perception of an LLM, \textit{is predicting missing parts of an image a better unsupervised learning signal than predicting the entire image representation produced by an external vision encoder?} According to Table~\ref{tab:competitors}, the answer is yes, as neither VIRAL nor the baseline model without masking can improve the original LLaVA on the Vision-Centric benchmarks. Notably, this holds true across both LLMs. 

Conversely, \ours with Vicuna-7B surpasses the standard LLaVA model on most visual tasks, recording an average gain of $+0.8$ points. Switching to Qwen2-7B, \ours outperforms LLaVA by an average margin of $+0.7$ points, with a significant improvement of $+6.2$ points on the challenging CVBench3D, while maintaining competitive or better scores on General, Knowledge, and OCR tasks. Some qualitative results are shown in Fig.~\ref{fig:qualitatives}, where we compare LLaVA, VIRAL, and \ours across multiple vision tasks, including CVBench2D, CVBench3D, and Blink.

\begin{table*}[t]
\centering
\caption{Performance of \ours evaluated across multiple LLM families.}
\vspace{-0.15cm}
\setlength{\tabcolsep}{.21em}
\resizebox{\linewidth}{!}{%
\begin{tabular}{lccc c c c c c c >{\columncolor{ourrow!30}}c>{\columncolor{ourrow!30}}c>{\columncolor{ourrow!30}}c>{\columncolor{ourrow!30}}c>{\columncolor{ourrow!30}}c>{\columncolor{ourrow!30}}c>{\columncolor{ourrow!30}}c>{\columncolor{ourrow!30}}c}
\toprule 
& & & & \textbf{General} & & \textbf{Knowledge} & & \textbf{OCR} & & \multicolumn{7}{c}{\cellcolor{ourrow!30}\textbf{Vision-Centric}} \\
\cmidrule(lr){5-5} \cmidrule(lr){7-7} \cmidrule(lr){9-9} \cmidrule{11-17}
& \textbf{LLM} & \textbf{Visual Encoder} & & Avg & &  Avg & & Avg && RealWorldQA & MMVP & Blink & CVBench2D & CVBench3D & Avg & \\
\midrule
LLaVA~\cite{liu2024improved} & Gemma2-2B~\cite{team2024gemma} & CLIP ViT-L & & \textbf{66.5} & & \textbf{42.0} & & \textbf{33.9} & & 53.3 & 44.3 & 30.7 & \textbf{57.3} & 64.7 & 50.0 & \\
\rowcolor{ourrow} 
\textbf{\ours (Ours)} &  & & & 66.1 & & 41.9 & & 33.4 & & \textbf{54.2} & \textbf{45.3} & \textbf{31.3} & \textbf{57.3} & \textbf{65.7} & \textbf{50.8} & \inc{0.8} \\
\midrule
LLaVA~\cite{liu2024improved} & LLaMA-3.2-3B~\cite{grattafiori2024llama} & CLIP ViT-L & & \textbf{67.4} & & \textbf{44.7} & & \textbf{34.7} & & 56.3 & 46.4 & \textbf{32.0} & \textbf{63.7} & 62.3 & 52.1 & \\
\rowcolor{ourrow} 
\textbf{\ours (Ours)} &  & & & 67.0 & & 44.3 & & 34.3 & & \textbf{58.7} & \textbf{49.2} & \textbf{32.0} & 63.6 & \textbf{64.1} & \textbf{53.5} & \inc{1.4} \\
\midrule
LLaVA~\cite{liu2024improved} & Vicuna-7B~\cite{vicuna2023} & CLIP ViT-L & & \textbf{65.8} & & 41.4 & & \textbf{36.2} & & 54.9 & \textbf{31.3} & 46.8 & 57.6 & 63.7 & 50.9 & \\
\rowcolor{ourrow}
\textbf{\ours (Ours)} &  & & & 65.1 & & \textbf{41.6} & & \textbf{36.2} & & \textbf{55.6} & 30.7 & \textbf{48.3} & \textbf{60.4} & \textbf{63.8} & \textbf{51.7} & \inc{0.8}\\
\midrule
LLaVA~\cite{liu2024improved} & Ministral-8B & CLIP ViT-L & & 65.5 & & \textbf{42.2} & & 33.0 & & \textbf{59.2} & 30.0 & 44.8 & \textbf{61.4} & 62.1 & 51.5 & \\
\rowcolor{ourrow}
\textbf{\ours (Ours)} &  & & & \textbf{67.2} & & \textbf{42.2} & & \textbf{34.0} & & 58.6 & \textbf{34.7} & \textbf{49.2} & 57.6 & \textbf{65.5} & \textbf{53.1} & \inc{1.6} \\
\midrule
LLaVA~\cite{liu2024improved} & Qwen2-7B~\cite{team2024qwen2} & CLIP ViT-L & & \textbf{72.0} & & 46.7 & & \textbf{36.9} & & \textbf{58.0} & 36.7 & \textbf{52.7} & 62.5 & 66.8 & 55.3 & \\
\rowcolor{ourrow}
\textbf{\ours (Ours)} &  & & &  71.9 & & \textbf{47.7} & & 36.3 & & 55.8 & \textbf{38.0} & 49.3 & \textbf{63.9} & \textbf{73.0} & \textbf{56.0} & \inc{0.7} \\
\midrule
LLaVA~\cite{liu2024improved} & Qwen2-7B~\cite{team2024qwen2} & SigLIP2 ViT-L & & 72.4 & & 47.4 & & 39.0 & & 57.5 & 48.5 & 48.0 & 65.8 & \textbf{68.1} & 57.6 & \\
\rowcolor{ourrow}
\textbf{\ours (Ours)} & & & & \textbf{74.0} & & \textbf{48.2} & & \textbf{43.8} & & \textbf{59.6} & \textbf{53.1} & \textbf{50.0} & \textbf{67.4} & 66.8 & \textbf{59.4} & \inc{1.8} \\
\bottomrule
\end{tabular}
}
\label{tab:comparison}
\vspace{-0.3cm}
\end{table*}

\tit{Generalization Across Different LLMs}
Table~\ref{tab:comparison} provides a comprehensive overview of the behavior of \ours across different LLM families\footnote{We refer to the supplementary material for additional details about the configurations used in these experiments.}. Beginning with the more compact language models, \ours consistently demonstrates improvements across visual tasks. For the Gemma2-2B~\cite{team2024gemma} model, \ours yields a $+0.8$ points gain in average Vision-Centric accuracy, elevating its performance from $50.0$ to $50.8$. Moving to the LLaMA-3.2-3B~\cite{grattafiori2024llama} model, \ours achieves an even more substantial $+1.4$ point increase, with the Vision-Centric average rising from $52.1$ to $53.5$. These results prove that \ours is effective in enhancing the fundamental visual understanding capabilities of even the more resource-constrained MLLMs. 

Scaling up the LLM, Vicuna-7B~\cite{vicuna2023} paired with \ours beats LLaVA on all visual tasks but MMVP, with a significant $+2.8$ gain on the challenging CVBench2D benchmark. With Ministral-8B, a variant of Mistral~\cite{jiang2023mistral7b}, we notice that \ours achieves superior performance on MMVP, Blink, and CVBench3D, reaching a $+1.6$ points gain in average accuracy on Vision-Centric tasks. Remarkably, \ours outscores LLaVA also on General tasks, moving from $65.5$ to $67.2$ points ($+1.7$), and from $33.0$ to $34.0$ points on OCR, while delivering the same accuracy as LLaVA on Knowledge. The comparison between \ours and LLaVA on Qwen2-7B, as previously discussed, confirms this trend, with \ours excelling on visual perception tasks such as CVBench2D and CVBench3D, while performing on par, or improving, over the other task categories. 

Overall, \ours when paired with Qwen2-7B delivers the best results among the considered LLMs. With that in mind, we also test this combination while scaling up the context visual encoder, switching from CLIP ViT-L/13@336~\cite{radford2021learning} to the more advanced SigLIP2-So400M L/14@384~\cite{tschannen2025siglip}. In this setting, \ours surpasses LLaVA on all four categories, exceeding LLaVA by $+1.6$ points on General, $+0.8$ points on Knowledge, $+4.8$ points on OCR, and $+1.8$ on Vision-Centric tasks. These results indicate that \ours benefits more from stronger visual encoders than the original visual instruction tuning~\cite{liu2024improved} recipe.

\subsection{Scaling the Target Encoder}
\begin{figure}[t]
    \centering
    \includegraphics[width=0.98\linewidth]{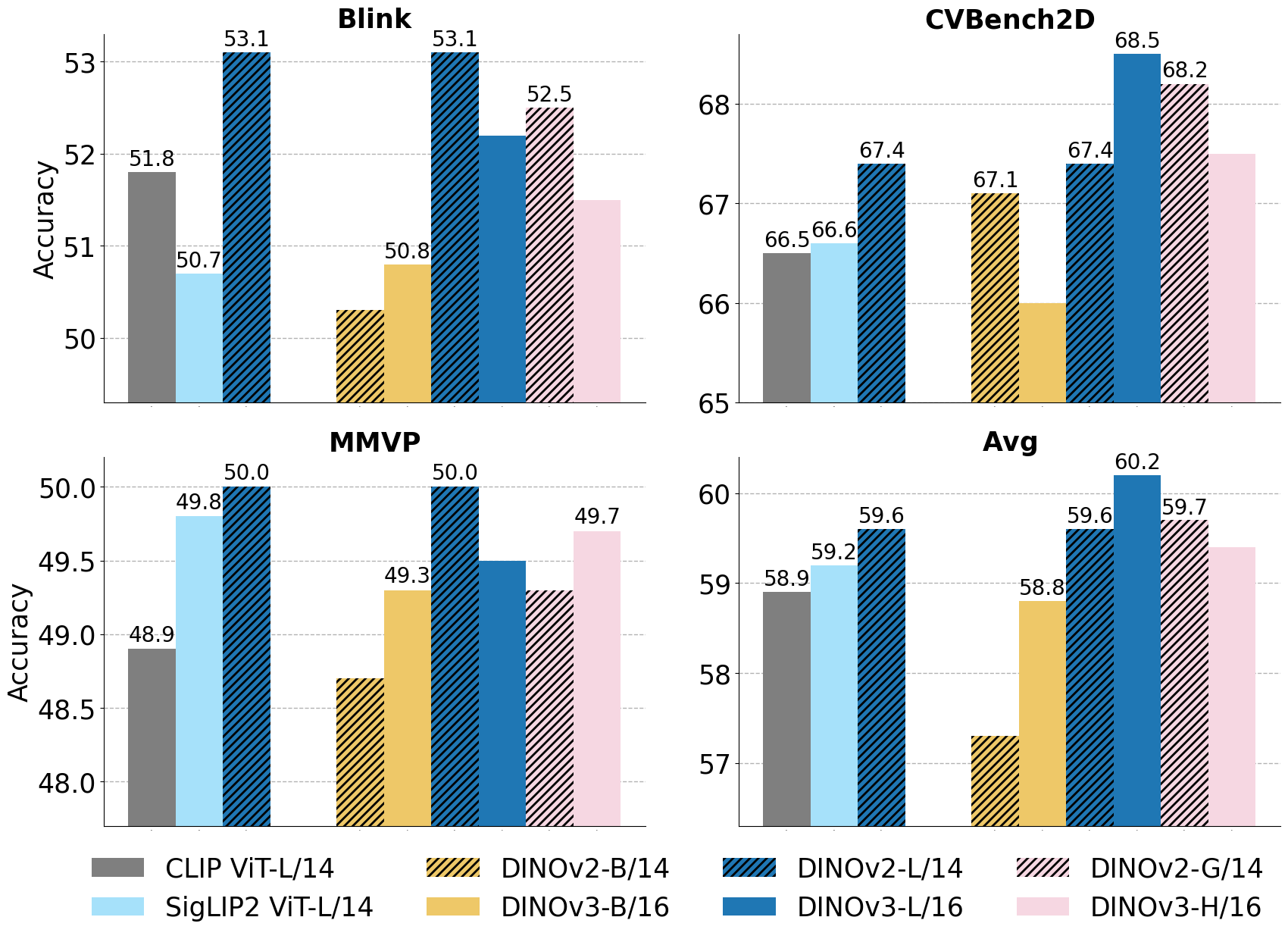}
    \vspace{-0.15cm}
    \caption{Effect of scaling the target visual encoder on Vision-Centric benchmarks.}
    \vspace{-0.35cm}
    \label{fig:graph}
\end{figure}

Proceeding with the best configuration of \ours, \ie~Qwen2-7B paired with SigLIP2 as the context visual encoder, we evaluate the impact of varying and scaling the target encoder. Table~\ref{fig:graph} reports \ours performance across three Vision-Centric datasets: Blink, CVBench2D, MMVP, and the average among all visual tasks. We consider visual encoder trained along with language supervision, such as CLIP~\cite{radford2021learning} and, SigLIP2~\cite{tschannen2025siglip}, as well as purely unsupervised models from the DINOv2~\cite{oquab2024dinov2} and DINOv3~\cite{simeoni2025dinov3} families, ranging from base to giant (1B) scales. All visual encoders works at resolution $[384 \times 384$]. We first compare CLIP, SigLIP2, and DINOv2, observing that DINOv2 consistently achieves the best performance across all datasets, testifying that dense features from language-supervised encoders are inferior to those from DINO. Motivated by this, we further explore multiple DINO variants, including Base, Large, Giant, and Huge models. Overall, Large or more advanced encoders tend to yield higher performance, demonstrating the benefits of scaling the target encoder in \ours.

\section{Conclusion}
In this paper, we introduced \ours, a framework designed to enhance the visual perception capabilities of MLLMs by integrating a self-supervised learning objective inspired by I-JEPA. The key insight behind \ours is that MLLMs can acquire more robust, fine-grained visual representations by predicting missing parts of an image in latent space, going beyond the inherent limitations of supervision derived solely from textual captions. By incorporating the predictive objective of I-JEPA during the alignment stage of LLaVA, \ours guides the LLM to capture the intrinsic structural and semantic regularities of the visual world, fostering a deeper understanding of visual content. Extensive experiments across diverse LLM families demonstrate that this approach consistently yields significant improvements on a wide array of vision-centric benchmarks, emphasizing that self-supervised visual learning enables MLLMs to effectively \textit{see beyond words} and better reason about complex visual information.

\section*{Acknowledgments}
We acknowledge the CINECA award under the ISCRA initiative and the EuroHPC supercomputer LUMI, for the availability of high-performance computing resources. This work has been supported by the EU Horizon project ``ELLIOT -  European Large Open Multi-Modal Foundation Models For Robust Generalization On Arbitrary Data Streams'' (No. 101214398), by the EuroHPC JU project ``MINERVA'' (GA No. 101182737), and by the PNRR project ``ITSERR'' (CUP B53C22001770006) funded by the EU - NextGenerationEU.

{
    \small
    \bibliographystyle{ieeenat_fullname}
    \bibliography{bibliography}
}


%

\clearpage
\maketitlesupplementary
\appendix
\section{Additional Implementation Details}
\label{supp:details}

\tinytit{Training Details} For all experiments, we follow the training hyperparameters proposed in LLaVA-1.5~\cite{liu2024improved}, which are detailed in Table~\ref{tab:llava_hyperparams}. From a computational perspective, \ours only requires the additonal forward pass of the target visual encoder during the alignment stage, which is relatively light compared to the forward pass of the LLM, and does not require computing any gradient.

\tit{LLM Details} We experiment with the instruction-tuned version of open-source LLMs that are freely available on Hugging Face's Transformers. We provide the exact reference to each LLM in Table~\ref{tab:llms_hf}.

\tit{Masking Details - $\tgtblock$} The objective of I-JEPA~\cite{assran2023self} is to predict the latent representation of $k$ target blocks of image patches $\{\tgtblock_j\}_{j=1,\dots,k}$, given access to a visible block of context image patches $\ctxblock$. In practice, these blocks are implemented as sets of integers, where each item in a set correspond to the index of a patch in the input image. 

For each sample, \ours predicts the representation of four, possibly overlapping, target blocks (\ie, $k=4$). In I-JEPA, the predictor runs a dedicated forward pass for each target block, where the input to each forward is the concatenation of the context embeddings with the positional encoding corresponding to the masked patches. Because in \ours the predictor is an LLM, for an efficient implementation, we pack the context embeddings and the $k$ targets, implemented as a learnable embedding plus positional encoding, into the same input, \ie, $[\imgctx, \latenttgt]$ in Eq.~\ref{eq:loss_jepa}, allowing a single forward pass on the LLM. To replicate the I-JEPA isolation of each target block, we modify the attention mask so that tokens belonging to different target blocks cannot attend to each other (see Fig.~\ref{fig:double_placeholder}). Note that information leakage between target blocks is still possible in case of overlap, even though it does not harm performance, as testified in Table~\ref{tab:ablation_supp} (top and middle). The scale of each target block $\tgtblock_j$, \ie, the number of masked patches with respect the total number of patches, is uniformly sampled for each batch within the range $(0.15, 0.20)$, while the aspect ratio is in $(0.75, 1.5)$.

\begin{table}[t]
\centering
\caption{
Training hyperparameters for the first training stage (\ie, alignment), as well as the second stage (\ie, visual instruction tuning). We use the same values according to LLaVA 1.5~\cite{liu2024improved}
}
\vspace{-0.15cm}
\resizebox{0.72\linewidth}{!}{%
\begin{tabular}{lc >{\columncolor{ourrow!30}}c >{\columncolor{ourrow!30}}c}
\toprule 
& & \multicolumn{2}{c}{\cellcolor{ourrow!30}\textbf{Training}} \\
\cmidrule(lr){3-4}
\textbf{Hyperparameter} && Stage 1 & Stage 2 \\
\midrule 
Batch Size & & 256 & 256 \\
Learning Rate & & 1e-3 & 2e-5 \\
Learning Rate Schedule & & \multicolumn{2}{c}{\cellcolor{ourrow!30}cosine decay} \\
Learning Rat Warmup Ratio & & \multicolumn{2}{c}{\cellcolor{ourrow!30}0.03} \\
Weight Decay & & \multicolumn{2}{c}{\cellcolor{ourrow!30}0} \\
Epochs & & \multicolumn{2}{c}{\cellcolor{ourrow!30}1} \\
Optimizer & & \multicolumn{2}{c}{\cellcolor{ourrow!30}AdamW~\cite{loshchilovdecoupled}} \\
DeepSpeed Stage~\cite{rajbhandari2020zero} & & 2 & 3 \\
\bottomrule
\end{tabular}
}
\label{tab:llava_hyperparams}
\vspace{-0.2cm}
\end{table}

\begin{table}[t]
\centering
\caption{
Mapping between each LLM used in this work and its page on Hugging Face~\cite{wolf-etal-2020-transformers}.
}
\vspace{-0.15cm}
\resizebox{0.92\linewidth}{!}{%
\begin{tabular}{ll}
\toprule
\textbf{LLM} & \textbf{Hugging Face Page} \\
\midrule
Gemma2-2B~\cite{team2024gemma} &
{\small\texttt{\href{https://huggingface.co/google/gemma-2-2b-it}{google/gemma-2-2b-it}}} \\
LLaMA-3.2-3B~\cite{grattafiori2024llama} &
{\small\texttt{\href{https://huggingface.co/meta-llama/Llama-3.2-3B-Instruct}{meta-llama/Llama-3.2-3B-Instruct}}} \\
Vicuna-7B~\cite{vicuna2023} &
{\small\texttt{\href{https://huggingface.co/lmsys/vicuna-7b-v1.5}{lmsys/vicuna-7b-v1.5}}} \\
Ministral-8B &
{\small\texttt{\href{https://huggingface.co/mistralai/Ministral-8B-Instruct-2410}{mistralai/Ministral-8B-Instruct-2410}}} \\
Qwen2-7B~\cite{team2024qwen2} &
{\small\texttt{\href{https://huggingface.co/Qwen/Qwen2-7B-Instruct}{Qwen/Qwen2-7B-Instruct}}} \\
\bottomrule
\end{tabular}
}
\vspace{-0.3cm}
\label{tab:llms_hf}
\end{table}

\begin{table*}[t!]
\centering
\caption{Additional ablation study results. All experiments employ Vicuna-7B as the underlying LLM with CLIP ViT-L@336 as context visual encoder and DINOv2-L/14 as target visual encoder.}
\vspace{-0.15cm}
\setlength{\tabcolsep}{.4em}
\resizebox{0.88\linewidth}{!}{%
\begin{tabular}{lc c c c c c c >{\columncolor{ourrow!30}}c>{\columncolor{ourrow!30}}c>{\columncolor{ourrow!30}}c>{\columncolor{ourrow!30}}c>{\columncolor{ourrow!30}}c>{\columncolor{ourrow!30}}c>{\columncolor{ourrow!30}}c}
\toprule 
& & \textbf{General} & & \textbf{Knowledge} & & \textbf{OCR} & & \multicolumn{6}{c}{\cellcolor{ourrow!30}\textbf{Vision-Centric}} \\
\cmidrule(lr){3-3} \cmidrule(lr){5-5} \cmidrule(lr){7-7} \cmidrule{9-14}
& & Avg & &  Avg & & Avg && RealWorldQA & MMVP & Blink & CVBench2D & CVBench3D & Avg \\
\midrule 
LLaVA~\cite{liu2024improved} & & 65.8 & & 41.4 & & 36.2 & & 54.9 & 31.3 & 46.8 & 57.6 & 63.7 & 50.9 \\
\midrule
\rowcolor{gray!10}
\multicolumn{14}{l}{\textit{Preventing targets overlap}} \\
w/o overlap & & \textbf{65.2} & & \textbf{41.7} & & 35.9 & & 55.3 & 28.7 & 46.8 & 58.7 & 61.1 & 50.1 \\
\rowcolor{ourrow}
w/ overlap & & 65.1 & & 41.6 & & \textbf{36.2} & & \textbf{55.6} & \textbf{30.7} & \textbf{48.3} & \textbf{60.4} & \textbf{63.8} & \textbf{51.7} \\
\midrule
\rowcolor{gray!10}
\multicolumn{14}{l}{\textit{Effect of changing attention patterns}} \\
$\attntgttotgt$ & & 64.7 & & 41.4 & & 35.7 & & \textbf{56.5} & 29.3 & 45.0 & 59.9 & 61.8 & 50.5 \\
$\nattntgttotxt$ & & 65.0 & & \textbf{42.2} & & 35.7 & & 54.9 & 25.3 & 46.2 & 57.3 & 62.5 & 49.2 \\
\rowcolor{ourrow}
$\attnours$ & & \textbf{65.1} & & 41.6 & & \textbf{36.2} & & 55.6 & \textbf{30.7} & \textbf{48.3} & \textbf{60.4} & \textbf{63.8} & \textbf{51.7} \\
\midrule
\rowcolor{gray!10}
\multicolumn{14}{l}{\textit{Effect of changing $\lambda$}} \\
$\lambda=0.0$ & & 63.8 & & \textbf{41.9} & & 35.6 & & \textbf{58.3} & 28.7 & 47.4 & 57.1 & 60.9 & 50.5 \\
\rowcolor{ourrow}
$\lambda=0.2$ & & \textbf{65.1} & & 41.6 & & \textbf{36.2} & & 55.6 & 30.7 & \textbf{48.3} & \textbf{60.4} & \textbf{63.8} & \textbf{51.7} \\
$\lambda=0.5$ & & 64.6 & & \textbf{41.9} & & 36.0 & & 57.1 & \textbf{32.0} & 45.8 & 59.0 & 62.3 & 51.2 \\
\bottomrule
\end{tabular}
}
\label{tab:ablation_supp}
\vspace{-0.35cm}
\end{table*}

\begin{figure}[t]
    \centering
    \includegraphics[width=\linewidth]{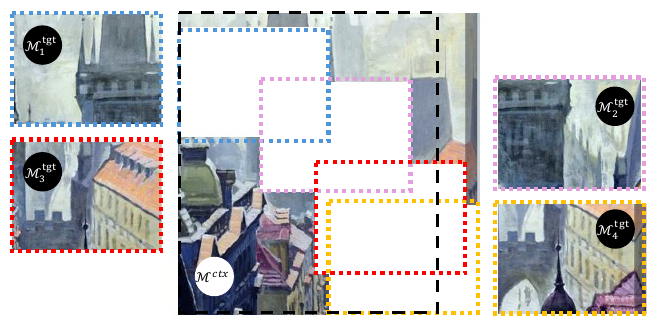}
    \vspace{-0.55cm}
    \caption{Example of the masking strategy of I-JEPA. First, four blocks of image patches $\{\tgtblock_j\}_{j=1,2,3,4}$ (\textit{colored, dotted} frames) are sampled with a given scale and aspect ratio. These will be processed by $\tgtvisenc$ and will be the prediction targets of $\lossjepa$. Concurrently, a larger block of patches $\ctxblock$ (\textit{dashed, black} frame) is sampled, and patches shared with the target blocks are removed. The remaining patches are converted into visual embeddings by $\ctxvisenc$, and will be forwarded to the LLM as a conditioning signal to predict the masked targets.}
    \label{fig:ctx_tgts_mask}
    \vspace{-0.2cm}
\end{figure}

\tit{Masking Details - $\ctxblock$} Similarly, the visible context $\ctxblock$ is a single block of patches with aspect ratio in $(0.75, 1.5)$, but which covers a larger part of the image. Indeed, the scale is uniformly sampled within $(0.85, 1.0)$. To prevent trivial predictions, we remove from $\ctxblock$ any index whose patch appears in any of the $k$ target blocks. An example of an input image being divided into context (\ie, \textit{dashed, black} frame) and target blocks (\ie, \textit{colored, dotted} frames) is depicted in Fig.~\ref{fig:ctx_tgts_mask}. We highlight that in practice the masking does not happen at pixel levels as the context visual encoder $\ctxvisenc$ still processes the whole, unmasked image. Conversely, the masking is applied on its output embeddings, so that only visual embeddings corresponding to $\ctxblock$ (\ie, $\imgctx$) are fed to the LLM, with the embeddings corresponding to target patches being replaced with the latent variable $\latenttgt$.

\section{Additional Experimental Results}
\label{supp:experiments}

\subsection{Additional Ablation Studies}

\tinytit{Validating the Masking Strategy} Table~\ref{tab:ablation_supp} (top) compares performance with and without overlap among target regions during training. When overlap between targets $\{\tgtblock_j\}$ is allowed, as in I-JEPA~\cite{assran2023self} (see Fig.~\ref{fig:ctx_tgts_mask}), the model achieves slightly higher scores across most benchmarks (\eg, $+1.6$ overall in the Vision-Centric average, from $50.1$ to $51.7$). This suggests that limited overlap can act as a mild regularizer, encouraging the model to better leverage spatial context and improving cross-region consistency. Conversely, enforcing non-overlapping targets (\textit{w/o overlap}) slightly reduces downstream performance, likely due to the reduced diversity and weaker spatial correspondence in the learned visual representations.

In Table~\ref{tab:ablation_supp} (middle), we further analyze the impact of different attention configurations between visual and textual tokens during training. We begin by discussing the choice of preventing tokens from different target blocks from attending to each other ($\nattntgttotgt$). We initially opt for this strategy to replicate I-JEPA, where the target blocks are predicted with different forward passes, and thus, there is no interaction between them. When we allow targets from different blocks to attend to each other, \ie, $\attntgttotgt$, performance remains stable on General, Knowledge, and OCR categories, but on Vision-Centric, it drops below the baseline level of LLaVA.

Next, we experiment by disabling attention from target tokens to textual tokens, \ie, $\nattntgttotxt$, thus precluding the LLM from looking at its own predictions for the missing parts of the image when generating the image description. This change effectively detaches the computation of $\lossntp$ from $\lossjepa$, and leads to a severe degradation in visual perception, with Vision-Centric average score dropping from $51.7$ to $49.2$. Overall, our proposed configuration, \ie, $\attnours$, which blocks attention between different target blocks while allowing text to attend to the predicted targets, achieves the best scores on General, OCR, and Vision-Centric tasks. 

\begin{table*}[t]
\centering
\caption{Ablation studies with Qwen2-7B as the underlying LLM, CLIP ViT-L@336 as context visual encoder, and DINOv2-L/14 as target visual encoder.}
\vspace{-0.15cm}
\setlength{\tabcolsep}{.35em}
\resizebox{\linewidth}{!}{%
\begin{tabular}{lc c c c c c c >{\columncolor{ourrow!30}}c>{\columncolor{ourrow!30}}c>{\columncolor{ourrow!30}}c>{\columncolor{ourrow!30}}c>{\columncolor{ourrow!30}}c>{\columncolor{ourrow!30}}c>{\columncolor{ourrow!30}}c}
\toprule 
& & \textbf{General} & & \textbf{Knowledge} & & \textbf{OCR} & & \multicolumn{6}{c}{\cellcolor{ourrow!30}\textbf{Vision-Centric}} \\
\cmidrule(lr){3-3} \cmidrule(lr){5-5} \cmidrule(lr){7-7} \cmidrule{9-14}
& & Avg & &  Avg & & Avg && RealWorldQA & MMVP & Blink & CVBench2D & CVBench3D & Avg \\
\midrule 
LLaVA~\cite{liu2024improved} & & 72.0 & & 46.7 & & 36.9 & & 58.0 & 36.7 & 52.7 & 62.5 & 66.8 & 55.3 \\
\midrule
\rowcolor{gray!10}
\multicolumn{14}{l}{\textit{Effect of changing the LLM layer}} \\
$j=4$  (early) & & \textbf{72.3} & & 47.4 & & 36.9 & & \textbf{57.5} & \textbf{39.3} & 47.7 & 63.7 & 69.2 & 55.5 \\
\rowcolor{ourrow}
$j=7$  (¼-depth) & & 71.9 & & \textbf{47.7} & & 36.3 & & 55.8 & 38.0 & 49.3 & \textbf{63.9} & \textbf{73.0} & \textbf{56.0} \\
$j=14$ (middle) & & 72.1 & & \textbf{47.7} & & 36.5 & & 56.5 & 34.7 & 48.9 & 63.0 & 69.4 & 54.5 \\
$j=21$ (¾-depth) & & 72.1 & & \textbf{47.7} & & 36.7 & & 56.9 & \textbf{39.3} & 48.9 & 62.9 & 66.5 & 54.9 \\
$j=27$ (near-final) & & 72.2 & & \textbf{47.7} & & \textbf{37.1} & & 56.5 & 33.3 & 48.9 & 62.4 & 66.3 & 53.5 \\
$j=28$ (final) & & 72.2 & & 47.6 & & 36.4 & & 56.2 & 37.3 & \textbf{50.2} & 62.5 & 67.3 & 54.7 \\
\midrule
\rowcolor{gray!10}
\multicolumn{14}{l}{\textit{Effect of projector scaling}} \\
w/ linear projection & & \textbf{72.0} & & \textbf{48.1} & & \textbf{36.6} & & \textbf{55.8} & \textbf{38.7} & 48.8 & 61.6 & 66.8 & 54.3 \\
\rowcolor{ourrow}
w/ MLP projection & & 71.9 & & 47.7 & & 36.3 & & \textbf{55.8} & 38.0 & \textbf{49.3} & \textbf{63.9} & \textbf{73.0} & \textbf{56.0}  \\
\midrule
\rowcolor{gray!10}
\multicolumn{14}{l}{\textit{Effect of changing the distance function in $\lossjepa$}} \\
w/ smooth L1 distance (I-JEPA~\cite{assran2023self}) & & 71.7 & & 47.6 & & \textbf{36.4} & & \textbf{57.0} & \textbf{40.0} & \textbf{49.8} & 62.0 & 66.3 & 55.0 \\
\rowcolor{ourrow}
w/ cosine distance & & \textbf{71.9} & & \textbf{47.7} & & 36.3 & & 55.8 & 38.0 & 49.3 & \textbf{63.9} & \textbf{73.0} & \textbf{56.0} \\
\bottomrule
\end{tabular}
}
\label{tab:ablation_qwen}
\vspace{-0.3cm}
\end{table*}
 
\tit{Effect of $\lossjepa$ Dropout Rate} In Table~\ref{tab:ablation_supp} (bottom), we present an ablation investigating the impact of changing the probability $\lambda$ associated with the $\lossjepa$ loss. As detailed in the main paper, this \textit{dropout} mechanism is introduced to balance the objectives of masked image modeling (via $\lossjepa$) and maintaining robust image-text alignment for subsequent instruction tuning.
We hypothesize that a small, non-zero dropout rate ($\lambda$) allows the model to periodically receive whole-image supervision, thereby improving its overall image-text alignment and downstream performance.
To validate this, we conducted an ablation study by varying the probability $\lambda \in \{0.0, 0.2, 0.5\}$ while keeping all other training parameters constant. 
When $\lambda=0.0$, $\lossjepa$ is always active. This leads to a performance degradation across most benchmarks, suggesting that continuous masked supervision compromises high-level image-text understanding. While using $\lambda=0.5$, we obtain slightly better results in certain tasks, for instance on the CVBench2D we pass from $57.1$ to $59.0$. However, the best overall performance is achieved with 
$\lambda=0.2$.  This value, adopted as our final choice in \ours, demonstrates the benefit of dynamically balancing visual (masked) and textual (whole-image next-token prediction) supervision.

\tit{Ablation Studies on Qwen2-7B}
We revise the ablation studies done in Table~\ref{tab:ablation} of the main paper, this time replacing Vicuna-7B~\cite{vicuna2023} with Qwen2-7B~\cite{team2024qwen2} as the underlying LLM for \ours. At first, we discuss the choice of the layer $j$ on top of which $\lossjepa$ is computed (Table~\ref{tab:ablation_supp}, top). Similarly to Vicuna-7B, selecting $j$ among the second half of the layers is ineffective in enhancing Vision-Centric performance against LLaVA. Also with Qwen2-7B, the best layer to improve visual perception is at one fourth of the depth, confirming that our choice can generalize beyond Vicuna-7B.
Next, we validate the effect of switching from an MLP network as a single linear projection to implement $\tgtproj$. The results (Table~\ref{tab:ablation_supp}, bottom) confirm the importance of employing a non-linear operator to convert the output from the $j$-th layer of Qwen2 to the target embedding space of DINOv2~\cite{oquab2024dinov2}, with an average gain on visual tasks of $+1.7$ points compared to the linear projector.
Finally, we apply the smooth L1 distance as the distance measure for $\lossjepa$ (Table~\ref{tab:ablation_qwen}, bottom). While the cosine distance remains the strongest choice in terms of average Vision-Centric accuracy, we highlight that the improvement over the smooth L1 distance, that amounts to $+2.6$ points when the LLM is Vicuna-7B, it is reduced to $1.0$ point on Qwen2-7B. This suggests that more advanced LLMs can be more easily aligned with vision foundation models, and thus predicting the magnitude along with the direction of visual embeddings becomes feasible for them.
 
\subsection{Detailed Results on CVBench}

\begin{table}[t]
\centering
\caption{Performance breakdown on single tasks of CVBench~\cite{tong2024cambrian}}
\vspace{-0.15cm}
\setlength{\tabcolsep}{.15em}
\resizebox{\linewidth}{!}{
\begin{tabular}{lcc c c c c c c}
\toprule 
 & & \multicolumn{3}{c}{CVBench2D} & & \multicolumn{3}{c}{CVBench3D} \\
\cmidrule(lr){3-5} \cmidrule(lr){7-9}
 & & $\text{Count}^{\textbf{2D}}$ & $\text{Relation}^{\textbf{2D}}$ & Overall & & $\text{Depth}^{\textbf{3D}}$ & $\text{Distance}^{\textbf{3D}}$ & Overall \\
\midrule
\rowcolor{gray!10}
\multicolumn{9}{l}{\textit{Vicuna-7B~\cite{vicuna2023}}} \\
LLaVA~\cite{liu2024improved} & & 57.0 & 59.8 & 57.6 & & 71.8 & 55.5 & 63.7 \\
VIRAL~\cite{yoon2025visual} & & 53.2 & 60.2 & 55.7 & & 64.2 & \textbf{55.8} & 60.0 \\
Ours w/o Masking & & 56.2 & 60.8 & 57.7 & & 69.5 & 52.0 & 60.8 \\
\rowcolor{ourrow}
\textbf{\ours (Ours)} & & \textbf{57.9} & \textbf{64.6} & \textbf{60.4} & & \textbf{73.4} & 54.2 & \textbf{63.8} \\
\midrule
\rowcolor{gray!10}
\multicolumn{9}{l}{\textit{Qwen2-7B~\cite{team2024qwen2}}} \\
LLaVA~\cite{liu2024improved} && 52.9 & 76.0 & 62.5 & & 76.3 & 57.3 & 66.8 \\
VIRAL~\cite{yoon2025visual} & &  53.2 & 74.2 & 62.0 & & 74.0 & 50.5 & 62.3 \\
Ours w/o Masking & &  53.2 & 76.5 & 62.8 & & 78.3 & 56.8 & 67.6 \\
\rowcolor{ourrow}
\textbf{\ours (Ours)} & & \textbf{53.8} & \textbf{78.2} & \textbf{63.9} & & \textbf{81.7} & \textbf{64.3} & \textbf{73.0} \\
\bottomrule
\end{tabular}
}
\label{tab:cvbench}
\vspace{-0.3cm}
\end{table}

\begin{table*}[t]
\centering
\caption{Results across General, Knowledge, OCR, and vision-centric benchmarks from the Cambrian Benchmark \cite{tong2024cambrian}.}
\vspace{-0.15cm}
\setlength{\tabcolsep}{.35em}
\resizebox{\linewidth}{!}{%
\begin{tabular}{lc c c c c c c c c c c c c c >{\columncolor{ourrow!30}}c>{\columncolor{ourrow!30}}c}
\toprule
& & \multicolumn{4}{c}{\textbf{General}} & & \multicolumn{4}{c}{\textbf{Knowledge}} &&  \multicolumn{3}{c}{\textbf{OCR}} && \multicolumn{1}{c}{\cellcolor{ourrow!30}\textbf{Vision-Centric}} \\
\cmidrule(lr){3-7} \cmidrule(lr){8-12} \cmidrule(lr){13-16} \cmidrule(lr){17-17}
& & MME & GQA & MMB & SEED && SQA & MMMU & MathVISTA & AI2D && ChartQA & OCRBench & TextVQA && Avg \\
\midrule
\rowcolor{gray!10}
\multicolumn{17}{l}{\textit{Vicuna-7B~\cite{vicuna2023} \& CLIP ViT-L}} \\
LLaVA~\cite{liu2024improved} & & \textbf{1507.4} & \textbf{62.6} & \textbf{62.1} & \textbf{66.8} && \textbf{69.6} & \textbf{35.1} & 5.7 & 55.2 && \textbf{17.8} & \textbf{32.}6 & \textbf{58.4} && 50.9  \\
\rowcolor{ourrow}
\textbf{\ours (Ours)} & & 1489.3 & 62.4 & 61.0 & \textbf{66.8} && 69.2 & 33.8 & \textbf{7.9} & \textbf{55.7} && 17.6 & \textbf{32.6} & 58.2 && \textbf{51.7}  \\
\midrule
\rowcolor{gray!10}
\multicolumn{17}{l}{\textit{Qwen2-7B~\cite{team2024qwen2} \& CLIP ViT-L}} \\
LLaVA~\cite{liu2024improved} & & 1603.7 & 63.0 & \textbf{73.3} & 70.3 && 74.8 & 39.4 & 7.1 & \textbf{65.7} && \textbf{19.1} & \textbf{32.8} & \textbf{58.7} && 55.3  \\
\rowcolor{ourrow}
\textbf{\ours (Ours)} & & \textbf{1585.5} & \textbf{63.3} & \textbf{73.3} & \textbf{70.4} && \textbf{76.3} & \textbf{41.8} & \textbf{7.9 }& 64.8 && 18.9 & 31.6 & 58.6 && \textbf{56.0}  \\
\midrule
\rowcolor{gray!10}
\multicolumn{17}{l}{\textit{Qwen2-7B~\cite{team2024qwen2} \& SigLIP2 ViT-L}} \\          
LLaVA~\cite{liu2024improved} & & 1575.5	& 63.9 & 73.7 & 71.7 && 75.2 & \textbf{43.2} & 6.9 & 64.2 && 17.8	& 37.1 & 62.2 && 57.6  \\
\rowcolor{ourrow}
\textbf{\ours (Ours)} & & \textbf{1612.2} & \textbf{64.9} & \textbf{75.4} & \textbf{73.5} && \textbf{76.5} & 42.0 & \textbf{8.5} & \textbf{66.0} && \textbf{25.8} & \textbf{40.5} & \textbf{65.1} && \textbf{59.4}  \\
\bottomrule
\end{tabular}
}
\label{tab:breakdown}
\vspace{-0.25cm}
\end{table*}

In Table~\ref{tab:cvbench}, we provide a detailed performance breakdown on the individual tasks of CVBench~\cite{tong2024cambrian}. \ours, consistently outperforms the LLaVA baseline and other methods across both the Vicuna-7B and Qwen2-7B LLMs. When using Vicuna-7B, \ours achieves the highest overall scores on both CVBench2D and CVBench3D, with a notable improvement in the $\text{Relation}^{\textbf{2D}}$ task, where it scores $64.6$ compared to the $59.8$ points from LLaVA. The performance advantages are even more pronounced with the stronger Qwen2-7B model, where \ours establishes the best results on all reported tasks. Specifically, it boosts the overall CVBench3D score to $73.0$, a $+6.2$ points increase over LLaVA, demonstrating substantial gains in both 3D depth (\ie, estimating distances from the camera) and distance understanding (\ie, estimating relative distances between objects). These results highlight the effectiveness of our approach on fine-grained, vision-centric evaluations.

\subsection{Complete Results on MLLM Tasks} 
Differently from the other experiments previously reported, Table~\ref{tab:breakdown} focuses on the General, Knowledge, and OCR categories, explicitly detailing the individual datasets included within the Cambrian evaluation benchmark~\cite{tong2024cambrian}. Specifically. as in the original paper, we group the datasets into three main categories more than the Vision-Centric one:

\begin{itemize}
    \item \textbf{General.} It includes MME~\cite{fu2023mme}, GQA~\cite{hudson2019gqa}, MMBench (MMB)~\cite{liu2023mmbench}, and SEED-Bench (SEED)~\cite{li2023seed}.
These datasets collectively evaluate perception and scene understanding through tasks such as quantification, color identification, and multi-domain visual comprehension.
    \item \textbf{Knowledge.} It comprises ScienceQA (SQA)~\cite{lu2022learn}, MMMU~\cite{yue2023mmmu}, MathVISTA~\cite{lu2023mathvista}, and AI2D~\cite{kembhavi2016diagram}.
This group measures factual and discipline-specific reasoning, testing models on science, mathematics, and diagram understanding that require textual and visual knowledge.
\item \textbf{OCR.}
It encompasses ChartQA~\cite{masry2022chartqa}, OCRBench~\cite{liu2024improved}, and TextVQA~\cite{singh2019towards}.
These benchmarks focus on recognizing and reasoning over embedded text and numerical information in images, assessing OCR accuracy and text-grounded visual reasoning.
\end{itemize}

As a key claim of this work, \ours strengthens visual reasoning while maintaining competitive performance on language- and knowledge-oriented tasks.
Notably, with Vicuna-7B, \ours achieves comparable results to LLaVA across general benchmarks and knowledge tasks, while preserving strong OCR accuracy. For example, although the baseline slightly outperforms by only $+0.2$ on GQA, our model matches its performance exactly on OCRBench.
When scaled to Qwen2-7B, using either CLIP ViT-L or SigLIP2 ViT-L as visual encoders, \ours consistently improves knowledge-intensive results (e.g., $+2.4$ on MMMU and $+0.8$ on MathVISTA with CLIP) while maintaining parity -- or even surpassing baselines -- on general and OCR tasks. Notably, on TextVQA with SigLIP2, performance increases from $62.2$ to $65.1$

\section{Additional Qualitative Results}
\label{supp:qualitatives}

\begin{figure*}[t!]
\begin{minipage}{0.24\linewidth}
\scriptsize{\textbf{Q}: Is this artwork titled the adoration\\ of the shepherds?
\vspace{0.15cm}}\\
\begin{minipage}{0.443\linewidth}
\includegraphics[width=1.\linewidth,height=0.88\linewidth]{images/qualitatives_supp/mme_45.png}
\end{minipage}
\hfill
\begin{minipage}{0.53\linewidth}
\scriptsize{
\textbf{LLaVA~\cite{liu2024improved}}:\\
Yes \textcolor{red}{\xmark} \\
\textbf{VIRAL~\cite{yoon2025visual}}:\\
Yes \textcolor{red}{\xmark} \\
\textbf{\ours (Ours):}\\
No \textcolor[HTML]{00b050}{\cmark}
}
\end{minipage}
\end{minipage}
\hspace{0.01cm}
\begin{minipage}{0.24\linewidth}
\scriptsize{\textbf{Q}: Is the person inside the red bounding box named Nick Porrazzo?
\vspace{0.15cm}}\\
\begin{minipage}{0.443\linewidth}
\includegraphics[width=1.\linewidth,height=0.88\linewidth]{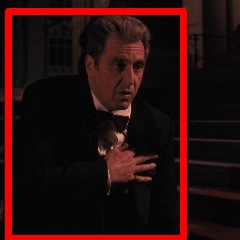}
\end{minipage}
\hfill
\begin{minipage}{0.53\linewidth}
\scriptsize{
\textbf{LLaVA~\cite{liu2024improved}}:\\
Yes \textcolor{red}{\xmark} \\
\textbf{VIRAL~\cite{yoon2025visual}}:\\
Yes \textcolor{red}{\xmark} \\
\textbf{\ours (Ours):}\\
No \textcolor[HTML]{00b050}{\cmark}
}
\end{minipage}
\end{minipage}
\hspace{0.01cm}
\begin{minipage}{0.24\linewidth}
\scriptsize{\textbf{Q}: Who is wearing a coat?\\
\vspace{0.15cm}}\\
\begin{minipage}{0.443\linewidth}
\includegraphics[width=1.\linewidth,height=0.88\linewidth]{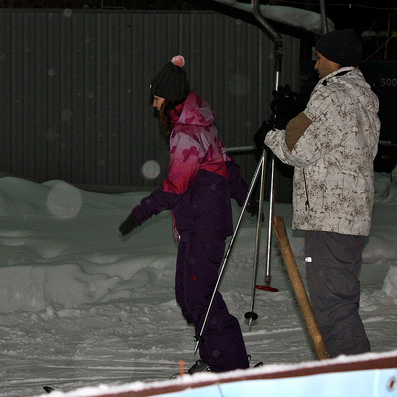}
\end{minipage}
\hfill
\begin{minipage}{0.53\linewidth}
\scriptsize{
\textbf{LLaVA~\cite{liu2024improved}}:\\
Man \textcolor[HTML]{00b050}{\cmark} \\
\textbf{VIRAL~\cite{yoon2025visual}}:\\
Girl \textcolor{red}{\xmark} \\
\textbf{\ours (Ours):}\\
Man \textcolor[HTML]{00b050}{\cmark}
}
\end{minipage}
\end{minipage}
\hspace{0.01cm}
%
\begin{minipage}{0.24\linewidth}
\scriptsize{\textbf{Q}: What can be found beyond the barbed wire fence?
\vspace{0.15cm}}\\
\begin{minipage}{0.443\linewidth}
\includegraphics[width=1.\linewidth,height=0.88\linewidth]{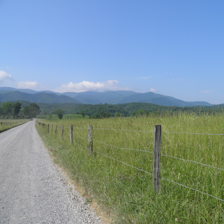}
\end{minipage}
\hfill
\begin{minipage}{0.53\linewidth}
\scriptsize{
\textbf{LLaVA~\cite{liu2024improved}}:\\
C) A small stream \textcolor{red}{\xmark} \\
\textbf{VIRAL~\cite{yoon2025visual}}:\\
 B) The horizon \textcolor{red}{\xmark} \\
\textbf{\ours (Ours):}\\
A) Another field \textcolor[HTML]{00b050}{\cmark}
}
\end{minipage}
\end{minipage}
\vspace{0.5cm}

\begin{minipage}{0.24\linewidth}
\scriptsize{\textbf{Q}: How many items sold less than 5 units in at least one store?
\vspace{0.15cm}}\\
\begin{minipage}{0.443\linewidth}
\includegraphics[width=1.\linewidth,height=0.88\linewidth]{images/qualitatives_supp/mathvista_10.png}
\end{minipage}
\hfill
\begin{minipage}{0.53\linewidth}
\scriptsize{
\textbf{LLaVA~\cite{liu2024improved}}:\\
10 \textcolor{red}{\xmark} \\
\textbf{VIRAL~\cite{yoon2025visual}}:\\
10 \textcolor{red}{\xmark} \\
\textbf{\ours (Ours):}\\
2 \textcolor[HTML]{00b050}{\cmark}
}
\end{minipage}
\end{minipage}
\hspace{0.01cm}
\begin{minipage}{0.24\linewidth}
\scriptsize{\textbf{Q}: What was the rate of change between Thursday and Friday in shells per day?
\vspace{0.15cm}}\\
\begin{minipage}{0.443\linewidth}
\includegraphics[width=1.\linewidth,height=0.88\linewidth]{images/qualitatives_supp/mathvista_358.png}
\end{minipage}
\hfill
\begin{minipage}{0.53\linewidth}
\scriptsize{
\textbf{LLaVA~\cite{liu2024improved}}:\\
-3 \textcolor{red}{\xmark} \\
\textbf{VIRAL~\cite{yoon2025visual}}:\\
-3 \textcolor{red}{\xmark} \\
\textbf{\ours (Ours):}\\
-7 \textcolor[HTML]{00b050}{\cmark}
}
\end{minipage}
\end{minipage}
\hspace{0.01cm}
\begin{minipage}{0.24\linewidth}
\scriptsize{\textbf{Q}: The interior of the walls and dome\\ of the building were covered with what?
\vspace{0.15cm}}\\
\begin{minipage}{0.443\linewidth}
\includegraphics[width=1.\linewidth,height=0.88\linewidth]{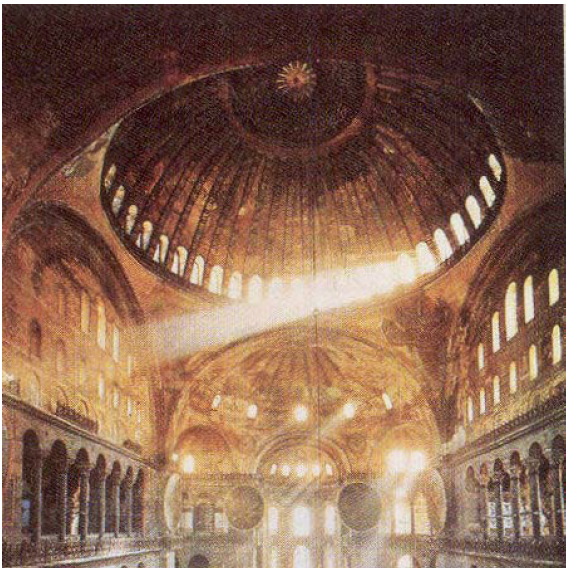}
\end{minipage}
\hfill
\begin{minipage}{0.53\linewidth}
\scriptsize{
\textbf{LLaVA~\cite{liu2024improved}}:\\
C) Mosaics \textcolor{red}{\xmark} \\
\textbf{VIRAL~\cite{yoon2025visual}}:\\
C) Mosaics  \textcolor{red}{\xmark} \\
\textbf{\ours (Ours):}\\
B) Frescoes \textcolor[HTML]{00b050}{\cmark}
}
\end{minipage}
\end{minipage}
\hspace{0.01cm}
%
\begin{minipage}{0.24\linewidth}
\scriptsize{\textbf{Q}: How many drive wheels are on the engine?
\vspace{0.15cm}}\\
\begin{minipage}{0.443\linewidth}
\includegraphics[width=1.\linewidth,height=0.88\linewidth]{images/qualitatives_supp/mmmu_715.png}
\end{minipage}
\hfill
\begin{minipage}{0.53\linewidth}
\scriptsize{
\textbf{LLaVA~\cite{liu2024improved}}:\\
D \textcolor{red}{\xmark} \\
\textbf{VIRAL~\cite{yoon2025visual}}:\\
C \textcolor{red}{\xmark} \\
\textbf{\ours (Ours):}\\
B \textcolor[HTML]{00b050}{\cmark}
}
\end{minipage}
\end{minipage}
\vspace{0.5cm}

\begin{minipage}{0.24\linewidth}
\scriptsize{\textbf{Q}: What's the least value of blue graph?\\
\vspace{0.15cm}}\\
\begin{minipage}{0.443\linewidth}
\includegraphics[width=1.\linewidth,height=0.88\linewidth]{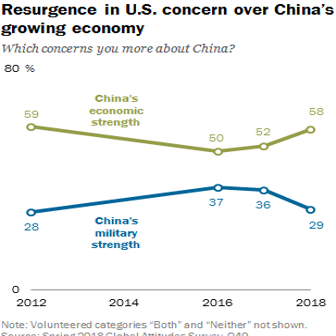}
\end{minipage}
\hfill
\begin{minipage}{0.53\linewidth}
\scriptsize{
\textbf{LLaVA~\cite{liu2024improved}}:\\
35 \textcolor{red}{\xmark} \\
\textbf{VIRAL~\cite{yoon2025visual}}:\\
25 \textcolor{red}{\xmark} \\
\textbf{\ours (Ours):}\\
28 \textcolor[HTML]{00b050}{\cmark}
}
\end{minipage}
\end{minipage}
\hspace{0.01cm}
\begin{minipage}{0.24\linewidth}
\scriptsize{\textbf{Q}: What \% support making voting\\ as easy as possible?
\vspace{0.15cm}}\\
\begin{minipage}{0.443\linewidth}
\includegraphics[width=1.\linewidth,height=0.88\linewidth]{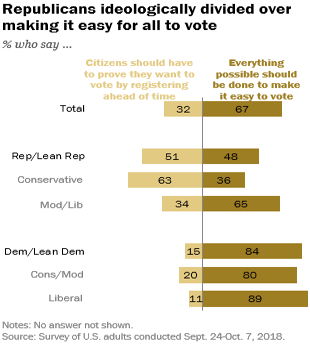}
\end{minipage}
\hfill
\begin{minipage}{0.53\linewidth}
\scriptsize{
\textbf{LLaVA~\cite{liu2024improved}}:\\
68 \textcolor{red}{\xmark} \\
\textbf{VIRAL~\cite{yoon2025visual}}:\\
76 \textcolor{red}{\xmark} \\
\textbf{\ours (Ours):}\\
67 \textcolor[HTML]{00b050}{\cmark}
}
\end{minipage}
\end{minipage}
\hspace{0.01cm}
\begin{minipage}{0.24\linewidth}
\scriptsize{\textbf{Q}: Which bank is hosting the cup?\\
\vspace{0.15cm}}\\
\begin{minipage}{0.443\linewidth}
\includegraphics[width=1.\linewidth,height=0.88\linewidth]{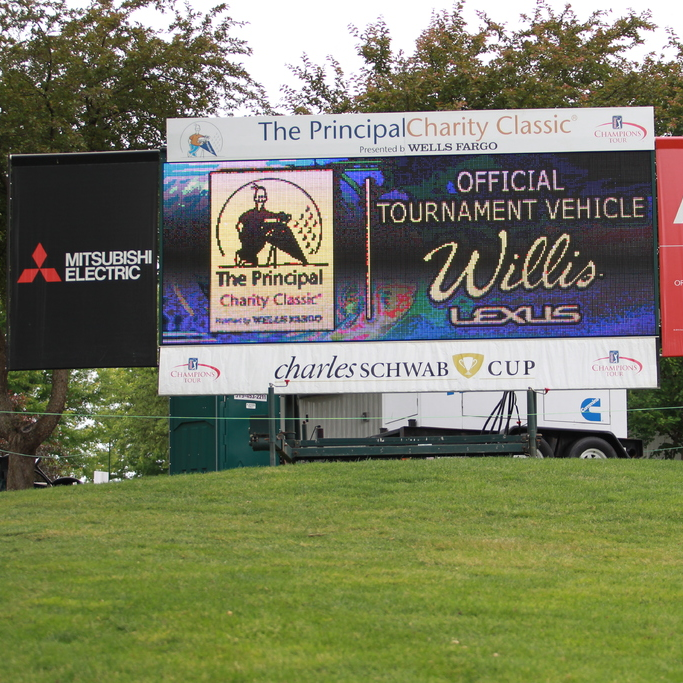}
\end{minipage}
\hfill
\begin{minipage}{0.53\linewidth}
\scriptsize{
\textbf{LLaVA~\cite{liu2024improved}}:\\
Wells fargo \textcolor{red}{\xmark} \\
\textbf{VIRAL~\cite{yoon2025visual}}:\\
Wells fargo  \textcolor{red}{\xmark} \\
\textbf{\ours (Ours):}\\
Charles schwab \textcolor[HTML]{00b050}{\cmark}
}
\end{minipage}
\end{minipage}
\hspace{0.01cm}
%
\begin{minipage}{0.24\linewidth}
\scriptsize{\textbf{Q}: How much is the compact space?\\
\vspace{0.15cm}}\\
\begin{minipage}{0.443\linewidth}
\includegraphics[width=1.\linewidth,height=0.88\linewidth]{images/qualitatives_supp/textvqa_2584.png}
\end{minipage}
\hfill
\begin{minipage}{0.53\linewidth}
\scriptsize{
\textbf{LLaVA~\cite{liu2024improved}}:\\
2299.5 \textcolor{red}{\xmark} \\
\textbf{VIRAL~\cite{yoon2025visual}}:\\
22995 \textcolor{red}{\xmark} \\
\textbf{\ours (Ours):}\\
229.95 \textcolor[HTML]{00b050}{\cmark}
}
\end{minipage}
\end{minipage}
\vspace{0.5cm}

\begin{minipage}{0.24\linewidth}
\scriptsize{\textbf{Q}: How many pedestrians are visible?\\
\vspace{0.15cm}}\\
\begin{minipage}{0.443\linewidth}
\includegraphics[width=1.\linewidth,height=0.88\linewidth]{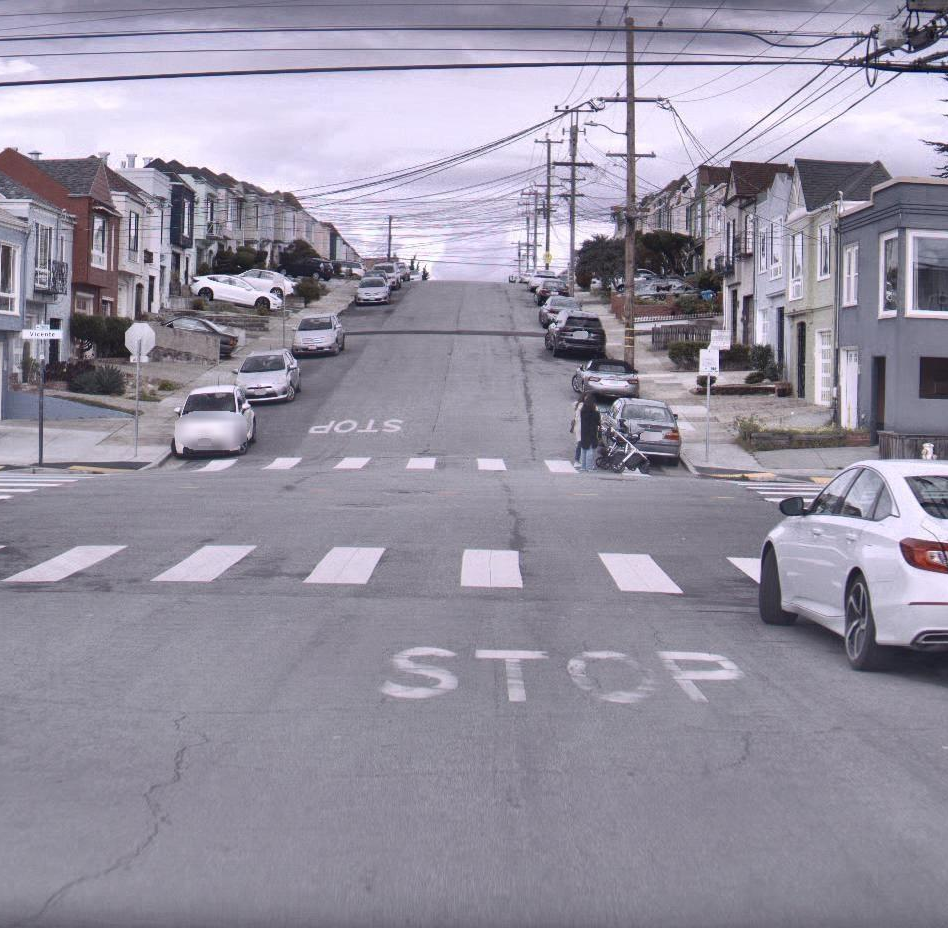}
\end{minipage}
\hfill
\begin{minipage}{0.53\linewidth}
\scriptsize{
\textbf{LLaVA~\cite{liu2024improved}}:\\
1 \textcolor{red}{\xmark} \\
\textbf{VIRAL~\cite{yoon2025visual}}:\\
1 \textcolor{red}{\xmark} \\
\textbf{\ours (Ours):}\\
2 \textcolor[HTML]{00b050}{\cmark}
}
\end{minipage}
\end{minipage}
\hspace{0.01cm}
\begin{minipage}{0.24\linewidth}
\scriptsize{\textbf{Q}: Where is the water?\\
\vspace{0.15cm}}\\
\begin{minipage}{0.443\linewidth}
\includegraphics[width=1.\linewidth,height=0.88\linewidth]{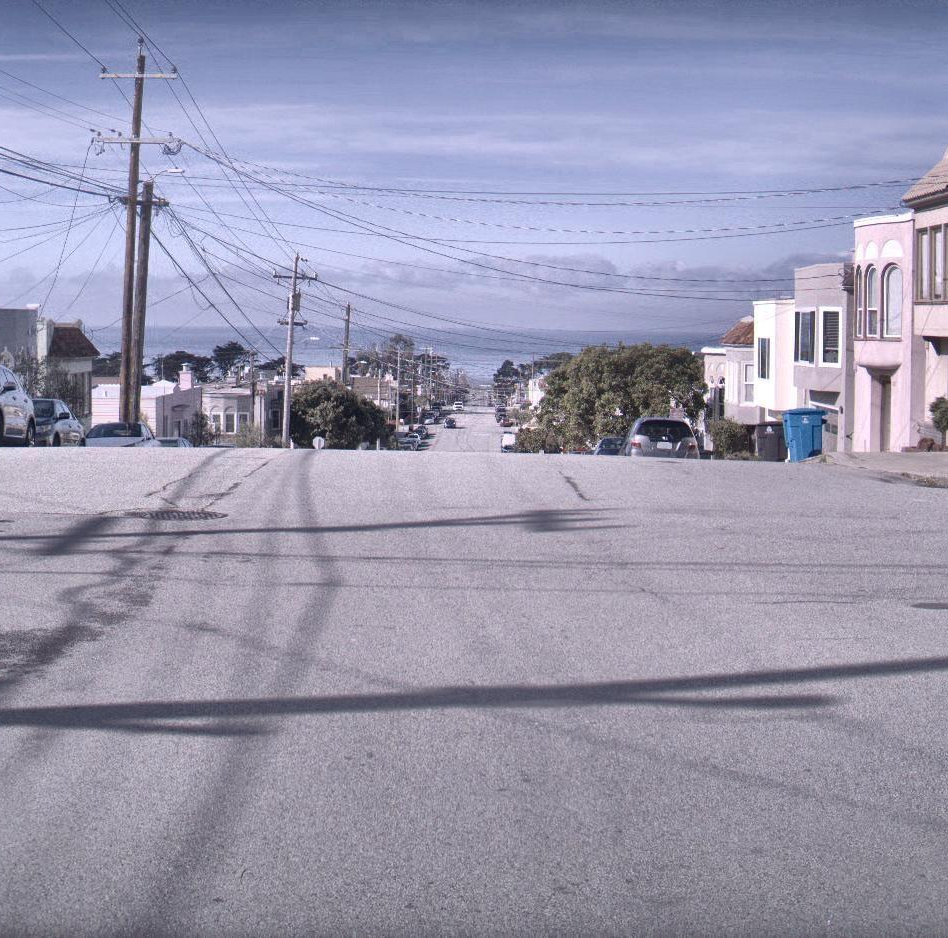}
\end{minipage}
\hfill
\begin{minipage}{0.53\linewidth}
\scriptsize{
\textbf{LLaVA~\cite{liu2024improved}}:\\
B) Behind you \textcolor{red}{\xmark} \\
\textbf{VIRAL~\cite{yoon2025visual}}:\\
C) To the right \textcolor{red}{\xmark} \\
\textbf{\ours (Ours):}\\
A) Straight ahead \textcolor[HTML]{00b050}{\cmark}
}
\end{minipage}
\end{minipage}
\hspace{0.01cm}
\begin{minipage}{0.24\linewidth}
\scriptsize{\textbf{Q}: Is the shark's belly visible\\in this image?
\vspace{0.15cm}}\\
\begin{minipage}{0.443\linewidth}
\includegraphics[width=1.\linewidth,height=0.88\linewidth]{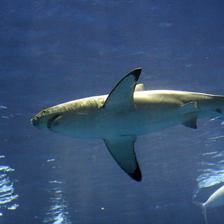}
\end{minipage}
\hfill
\begin{minipage}{0.53\linewidth}
\scriptsize{
\textbf{LLaVA~\cite{liu2024improved}}:\\
B) No \textcolor{red}{\xmark} \\
\textbf{VIRAL~\cite{yoon2025visual}}:\\
B) No  \textcolor{red}{\xmark} \\
\textbf{\ours (Ours):}\\
A) Yes \textcolor[HTML]{00b050}{\cmark}
}
\end{minipage}
\end{minipage}
\hspace{0.01cm}
%
\begin{minipage}{0.24\linewidth}
\scriptsize{\textbf{Q}: Are there cookies stacked on top of other cookies?
\vspace{0.15cm}}\\
\begin{minipage}{0.443\linewidth}
\includegraphics[width=1.\linewidth,height=0.88\linewidth]{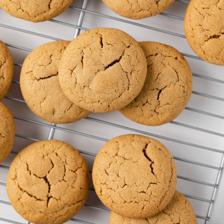}
\end{minipage}
\hfill
\begin{minipage}{0.53\linewidth}
\scriptsize{
\textbf{LLaVA~\cite{liu2024improved}}:\\
B) No \textcolor{red}{\xmark} \\
\textbf{VIRAL~\cite{yoon2025visual}}:\\
B) No \textcolor{red}{\xmark} \\
\textbf{\ours (Ours):}\\
A) Yes \textcolor[HTML]{00b050}{\cmark}
}
\end{minipage}
\end{minipage}
\vspace{0.5cm}

\begin{minipage}{0.24\linewidth}
\scriptsize{\textbf{Q}: Which bounding box more accurately localizes and encloses the train?
\vspace{0.15cm}}\\
\begin{minipage}{0.443\linewidth}
\includegraphics[width=1.\linewidth,height=0.88\linewidth]{images/qualitatives_supp/blink_loc_29.png}
\end{minipage}
\hfill
\begin{minipage}{0.53\linewidth}
\scriptsize{
\textbf{LLaVA~\cite{liu2024improved}}:\\
A) Box A \textcolor{red}{\xmark} \\
\textbf{VIRAL~\cite{yoon2025visual}}:\\
A) Box A \textcolor{red}{\xmark} \\
\textbf{\ours (Ours):}\\
B) Box B \textcolor[HTML]{00b050}{\cmark}
}
\end{minipage}
\end{minipage}
\hspace{0.01cm}
\begin{minipage}{0.24\linewidth}
\scriptsize{\textbf{Q}: Consider the surface color of the two points A and B. Which one is darker?
\vspace{0.15cm}}\\
\begin{minipage}{0.443\linewidth}
\includegraphics[width=1.\linewidth,height=0.88\linewidth]{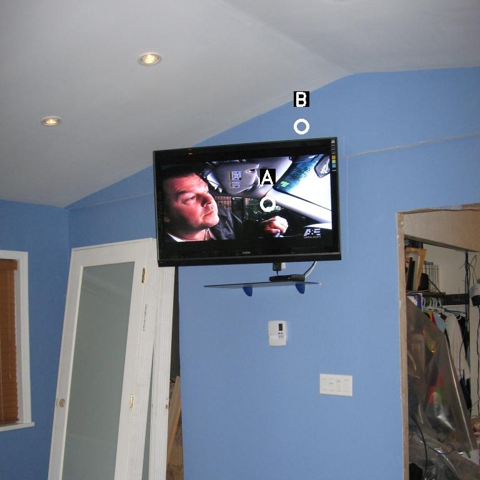}
\end{minipage}
\hfill
\begin{minipage}{0.53\linewidth}
\scriptsize{
\textbf{LLaVA~\cite{liu2024improved}}:\\
C) Same color \textcolor{red}{\xmark} \\
\textbf{VIRAL~\cite{yoon2025visual}}:\\
C) Same color \textcolor{red}{\xmark} \\
\textbf{\ours (Ours):}\\
A) A is darker \textcolor[HTML]{00b050}{\cmark}
}
\end{minipage}
\end{minipage}
\hspace{0.01cm}
\begin{minipage}{0.24\linewidth}
\scriptsize{\textbf{Q}: Points A and B are circled on the image. Which one is closer to the camera?
\vspace{0.15cm}}\\
\begin{minipage}{0.443\linewidth}
\includegraphics[width=1.\linewidth,height=0.88\linewidth]{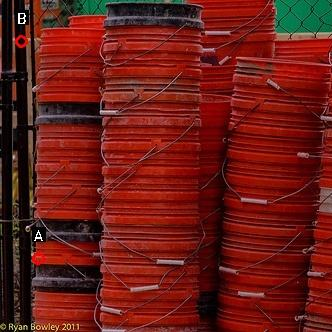}
\end{minipage}
\hfill
\begin{minipage}{0.53\linewidth}
\scriptsize{
\textbf{LLaVA~\cite{liu2024improved}}:\\
B) B is closer \textcolor{red}{\xmark} \\
\textbf{VIRAL~\cite{yoon2025visual}}:\\
B) B is closer  \textcolor{red}{\xmark} \\
\textbf{\ours (Ours):}\\
A) A is closer \textcolor[HTML]{00b050}{\cmark}
}
\end{minipage}
\end{minipage}
\hspace{0.01cm}
%
\begin{minipage}{0.24\linewidth}
\scriptsize{\textbf{Q}: Points A and B are circled on the image. Which one is closer to the camera?
\vspace{0.15cm}}\\
\begin{minipage}{0.443\linewidth}
\includegraphics[width=1.\linewidth,height=0.88\linewidth]{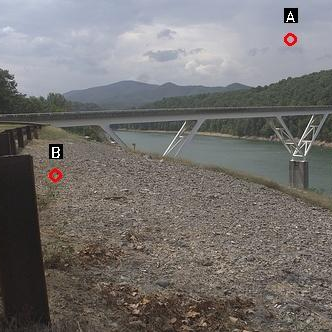}
\end{minipage}
\hfill
\begin{minipage}{0.53\linewidth}
\scriptsize{
\textbf{LLaVA~\cite{liu2024improved}}:\\
A) A is closer \textcolor{red}{\xmark} \\
\textbf{VIRAL~\cite{yoon2025visual}}:\\
A) A is closer \textcolor{red}{\xmark} \\
\textbf{\ours (Ours):}\\
B) B is closer \textcolor[HTML]{00b050}{\cmark}
}
\end{minipage}
\end{minipage}
\vspace{0.5cm}

%
\begin{minipage}{0.24\linewidth}
\scriptsize{\textbf{Q}: Where is the sconce (red box) located with respect to the picture?
\vspace{0.15cm}}\\
\begin{minipage}{0.443\linewidth}
\includegraphics[width=1.\linewidth,height=0.88\linewidth]{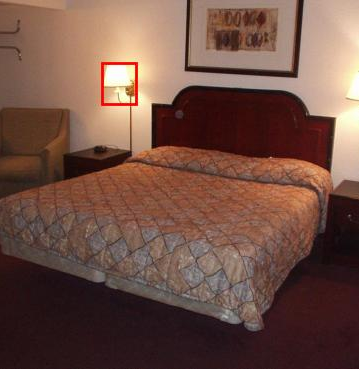}
\end{minipage}
\hfill
\begin{minipage}{0.53\linewidth}
\scriptsize{
\textbf{LLaVA~\cite{liu2024improved}}:\\
B) Right \textcolor{red}{\xmark} \\
\textbf{VIRAL~\cite{yoon2025visual}}:\\
B) Right \textcolor{red}{\xmark} \\
\textbf{\ours (Ours):}\\
A) Left \textcolor[HTML]{00b050}{\cmark}
}
\end{minipage}
\end{minipage}
\hspace{0.01cm}
\begin{minipage}{0.24\linewidth}
\scriptsize{\textbf{Q}: Where is the person located with\\ respect to the frisbee?
\vspace{0.15cm}}\\
\begin{minipage}{0.443\linewidth}
\includegraphics[width=1.\linewidth,height=0.88\linewidth]{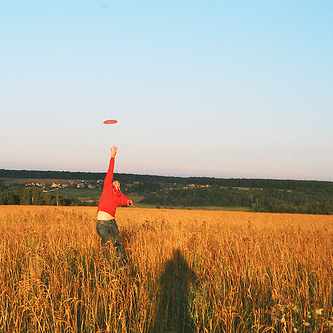}
\end{minipage}
\hfill
\begin{minipage}{0.53\linewidth}
\scriptsize{
\textbf{LLaVA~\cite{liu2024improved}}:\\
A) Above \textcolor{red}{\xmark} \\
\textbf{VIRAL~\cite{yoon2025visual}}:\\
A) Above \textcolor{red}{\xmark} \\
\textbf{\ours (Ours):}\\
B) Below \textcolor[HTML]{00b050}{\cmark}
}
\end{minipage}
\end{minipage}
\hspace{0.01cm}
\begin{minipage}{0.24\linewidth}
\scriptsize{\textbf{Q}: Which is closer to the camera, the towel (red box) or the lamp (blue box)?
\vspace{0.15cm}}\\
\begin{minipage}{0.443\linewidth}
\includegraphics[width=1.\linewidth,height=0.88\linewidth]{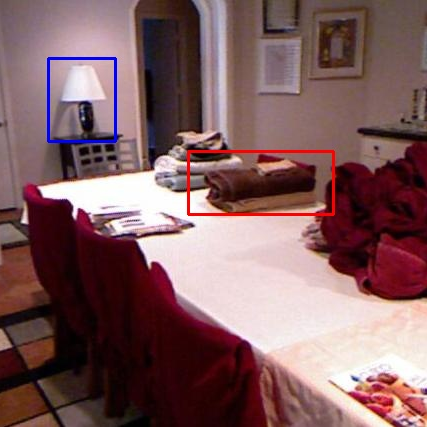}
\end{minipage}
\hfill
\begin{minipage}{0.53\linewidth}
\scriptsize{
\textbf{LLaVA~\cite{liu2024improved}}:\\
B) Lamp \textcolor{red}{\xmark} \\
\textbf{VIRAL~\cite{yoon2025visual}}:\\
B) Lamp  \textcolor{red}{\xmark} \\
\textbf{\ours (Ours):}\\
A) Towel \textcolor[HTML]{00b050}{\cmark}
}
\end{minipage}
\end{minipage}
\hspace{0.01cm}
\begin{minipage}{0.24\linewidth}
\scriptsize{\textbf{Q}: Which is closer to the car (red box), the bicycle (blue box) or the bus (green box)?
\vspace{0.15cm}}\\
\begin{minipage}{0.443\linewidth}
\includegraphics[width=1.\linewidth,height=0.88\linewidth]{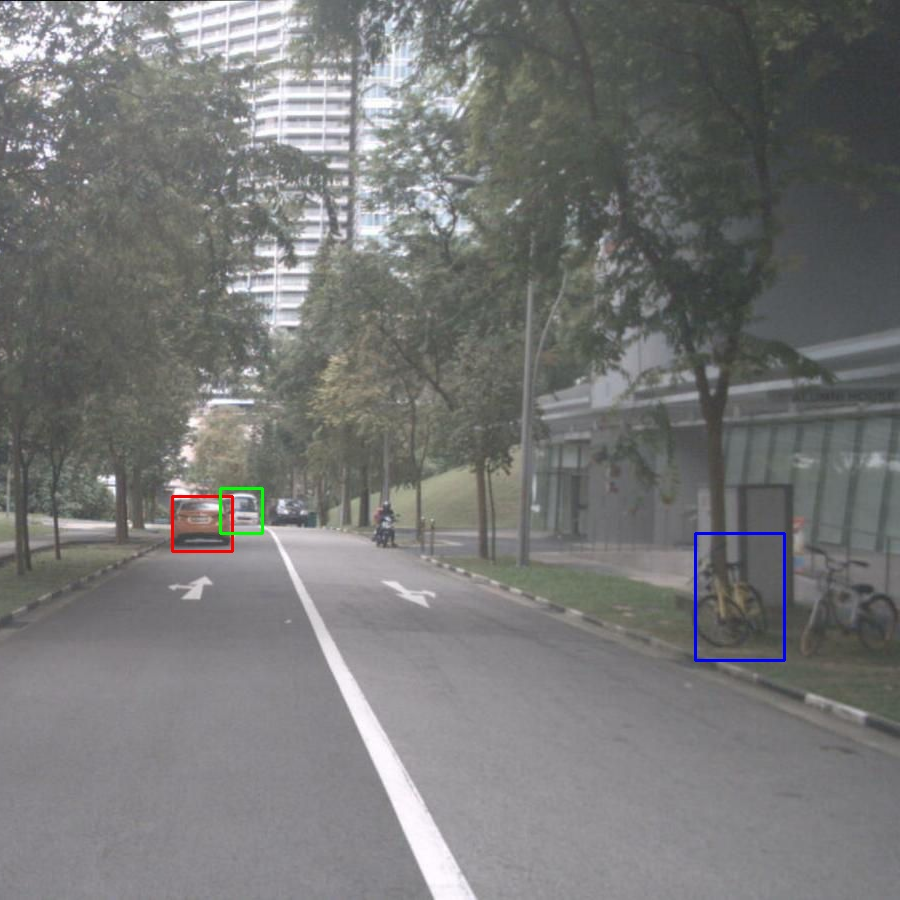}
\end{minipage}
\hfill
\begin{minipage}{0.53\linewidth}
\scriptsize{
\textbf{LLaVA~\cite{liu2024improved}}:\\
B) Bus \textcolor{red}{\xmark} \\
\textbf{VIRAL~\cite{yoon2025visual}}:\\
B) Bus \textcolor{red}{\xmark} \\
\textbf{\ours (Ours):}\\
A) Bicycle \textcolor[HTML]{00b050}{\cmark}
}
\end{minipage}
\end{minipage}
\vspace{-0.1cm}
\caption{Qualitative comparison of three training methods for MLLMs with Qwen2-7B~\cite{team2024qwen2}. We show samples from all categories of Cambrian~\cite{tong2024cambrian}: General (first row), Knowledge (second row), OCR (third row), and Vision-Centric (last three rows).}
\label{fig:qualitatives_supp}
\vspace{0.3cm}
\end{figure*}

In this section, we present additional qualitative examples showcasing the behavior of our \ours model across the four major categories of the Cambrian benchmark. Fig.~\ref{fig:qualitatives_supp} provides a comparison between LLaVA~\cite{liu2024improved}, VIRAL~\cite{yoon2025visual}, and our approach. The first row includes samples drawn from the General category, covering broad visual recognition and commonsense queries. The second row focuses on Knowledge-intensive examples, where models must rely on factual or domain-specific visual grounding. The third row evaluates OCR capabilities, involving charts, text-heavy images, or numeric reasoning. Finally, the last three rows include Vision-Centric cases.
Overall, \ours demonstrates strong visual grounding and geometric reasoning, correctly handling reflectance differences, depth ordering, counting, and occlusions, areas where prior MLLMs frequently fail. Beyond this, our method also consistently outperforms baselines and competitor in the other categories, showing more reliable recognition, more accurate reading of charts and graphs. These qualitative examples highlight the robustness and versatility of our approach across diverse visual reasoning scenarios.

\begin{figure*}[t!]
\begin{minipage}{0.24\linewidth}
\scriptsize{\textbf{Q}: Is the following statement correct:\\The spider has 8 legs?
\vspace{0.15cm}}\\
\begin{minipage}{0.443\linewidth}
\includegraphics[width=1.\linewidth,height=0.88\linewidth]{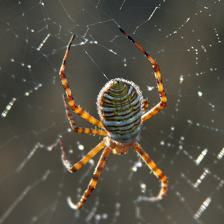}
\end{minipage}
\hfill
\begin{minipage}{0.53\linewidth}
\scriptsize{
\textbf{LLaVA~\cite{liu2024improved}}:\\
A) Yes \textcolor{red}{\xmark} \\
\textbf{VIRAL~\cite{yoon2025visual}}:\\
A) Yes \textcolor{red}{\xmark} \\
\textbf{\ours (Ours):}\\
A) Yes \textcolor{red}{\xmark}
}
\end{minipage}
\end{minipage}
\hspace{0.01cm}
\begin{minipage}{0.24\linewidth}
\scriptsize{\textbf{Q}: Is the decoration on the Easter egg flat or raised?
\vspace{0.15cm}}\\
\begin{minipage}{0.443\linewidth}
\includegraphics[width=1.\linewidth,height=0.88\linewidth]{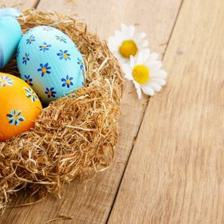}
\end{minipage}
\hfill
\begin{minipage}{0.53\linewidth}
\scriptsize{
\textbf{LLaVA~\cite{liu2024improved}}:\\
B) Raised \textcolor{red}{\xmark} \\
\textbf{VIRAL~\cite{yoon2025visual}}:\\
B) Raised \textcolor{red}{\xmark} \\
\textbf{\ours (Ours):}\\
B) Raised \textcolor{red}{\xmark}
}
\end{minipage}
\end{minipage}
\hspace{0.01cm}
\begin{minipage}{0.24\linewidth}
\scriptsize{\textbf{Q}: Where is the potted plant (red box) located with respect to the horse?
\vspace{0.15cm}}\\
\begin{minipage}{0.443\linewidth}
\includegraphics[width=1.\linewidth,height=0.88\linewidth]{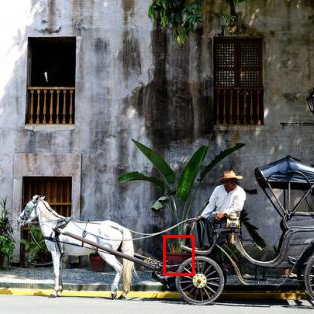}
\end{minipage}
\hfill
\begin{minipage}{0.53\linewidth}
\scriptsize{
\textbf{LLaVA~\cite{liu2024improved}}:\\
A) Left \textcolor{red}{\xmark} \\
\textbf{VIRAL~\cite{yoon2025visual}}:\\
A) Left \textcolor{red}{\xmark} \\
\textbf{\ours (Ours):}\\
A) Left \textcolor{red}{\xmark}
}
\end{minipage}
\end{minipage}
\hspace{0.01cm}
%
\begin{minipage}{0.24\linewidth}
\scriptsize{\textbf{Q}: Which is closer to the camera, the tissues (red box) or the bin (blue box)?
\vspace{0.15cm}}\\
\begin{minipage}{0.443\linewidth}
\includegraphics[width=1.\linewidth,height=0.88\linewidth]{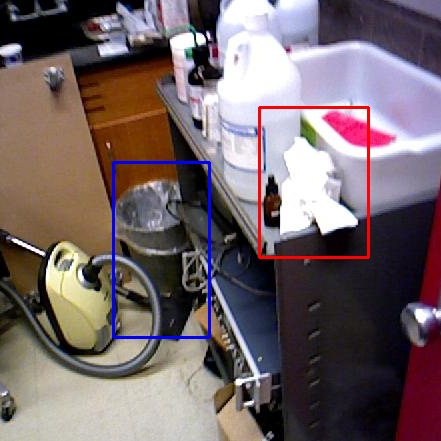}
\end{minipage}
\hfill
\begin{minipage}{0.53\linewidth}
\scriptsize{
\textbf{LLaVA~\cite{liu2024improved}}:\\
B) Bin \textcolor{red}{\xmark} \\
\textbf{VIRAL~\cite{yoon2025visual}}:\\
B) Bin \textcolor{red}{\xmark} \\
\textbf{\ours (Ours):}\\
B) Bin \textcolor{red}{\xmark}
}
\end{minipage}
\end{minipage}
\vspace{-0.1cm}
\caption{Some failure cases on Vision-Centric tasks employing Qwen2-7B~\cite{team2024qwen2}. We show samples from MMVP~\cite{tong2024eyes} (1st and 2nd columns), CVBench2D~\cite{tong2024cambrian} (3rd column), and CVBench3D~\cite{tong2024cambrian} (4th column).}
\label{fig:failures}
\vspace{-.3cm}
\end{figure*}

\section{Limitations and Impact}
\label{supp:limitations}
In this work, we have demonstrated both quantitatively and qualitatively that \ours enhances the visual perceptions in MLLMs through self-supervised visual learning. Yet, we acknowledge that much has been left to be done towards human-level spatial reasoning, as testified by the failure cases reported in Fig.~\ref{fig:failures}. For instance, we can see from the first example that all MLLMs confirm that the spider has eight legs, whereas it has only six. Moreover, we empirically find the best performing layer, \ie, at one fourth of the depth of the LLM, on top of which we compute $\lossjepa$. Despite experiments demonstrating that this choice transfers well across models, we hope that future works will bring a principled and theoretical criterion for selecting the best layer for self-supervised visual learning.

\end{document}